\documentclass[10pt,twocolumn,letterpaper]{article}

\usepackage[pagenumbers]{cvpr} 
\usepackage[accsupp]{axessibility}
\usepackage{algorithm}
\usepackage{algpseudocode}
\usepackage{newfloat}
\usepackage{listings}

\usepackage{amssymb}
\usepackage{booktabs}
\usepackage{soul}
\usepackage{multirow}
\usepackage{subcaption}
\usepackage{adjustbox}
\usepackage{makecell}

\usepackage{amsthm}

\theoremstyle{definition}

\theoremstyle{remark}

\usepackage{colortbl,xcolor} 
\definecolor{robblue}{RGB}{220,230,255}
\definecolor{robgreen}{RGB}{205,240,210}

\usepackage{pifont}

\renewcommand{\paragraph}[1]{
     \noindent{\textbf{#1}} 
 }

\usepackage[cjk]{kotex}

\usepackage{bm}

\definecolor{cvprblue}{rgb}{0.21,0.49,0.74}
\usepackage[pagebackref,breaklinks,colorlinks,allcolors=cvprblue]{hyperref}

\def\paperID{} 
\def\confName{CVPR}
\def\confYear{2026}

\title{When CLIP Sees More, It Fights Back Harder:
Multi-View Guided Adaptive Counterattacks for Test-Time Adversarial Robustness}

\author{Sunoh Kim\\
Dankook University\\
Yongin, South Korea\\
{\tt\small suno8386@dankook.ac.kr}
\and
Daeho Um\thanks{Corresponding author.}\\
University of Seoul\\
Seoul, South Korea\\
{\tt\small daehoum@uos.ac.kr}
}

\begin{document}
\maketitle
\begin{abstract}
Vision-language models such as CLIP have achieved remarkable zero-shot recognition capabilities, yet their robustness against adversarial perturbations remains limited. Test-time counterattack (TTC) was recently proposed to improve CLIP's robustness by perturbing an input image to steer it away from a corrupted state during inference. However, TTC remains fragile under strong attacks because its counterattack relies on a directly corrupted original view and employs a noise-driven hard-gating scheme that cannot adapt to varying corruption severity. To address these limitations, we introduce Multi-view guided Adaptive Counterattack (MAC), which performs counterattacks for multi-view with corruption-aware soft weighting. Specifically, MAC first constructs augmented views of an input image to obtain diverse embeddings. It then performs counterattacks to refine corrupted embeddings of views. Next, MAC adaptively scales the counterattack intensity for each view based on its estimated corruption degree. Finally, the adaptively counterattacked views are aggregated to yield a robust final prediction. Extensive experiments across 20 datasets and diverse attack scenarios demonstrate that MAC substantially improves robustness while preserving high inference speed and memory efficiency with its tuning-free design. Our code is available at \url{https://github.com/sunoh-kim/MAC}.
\end{abstract}    
\section{Introduction}
\label{sec:intro}

\begin{figure}[t!]
  \centering
  \includegraphics[width=0.93\linewidth]{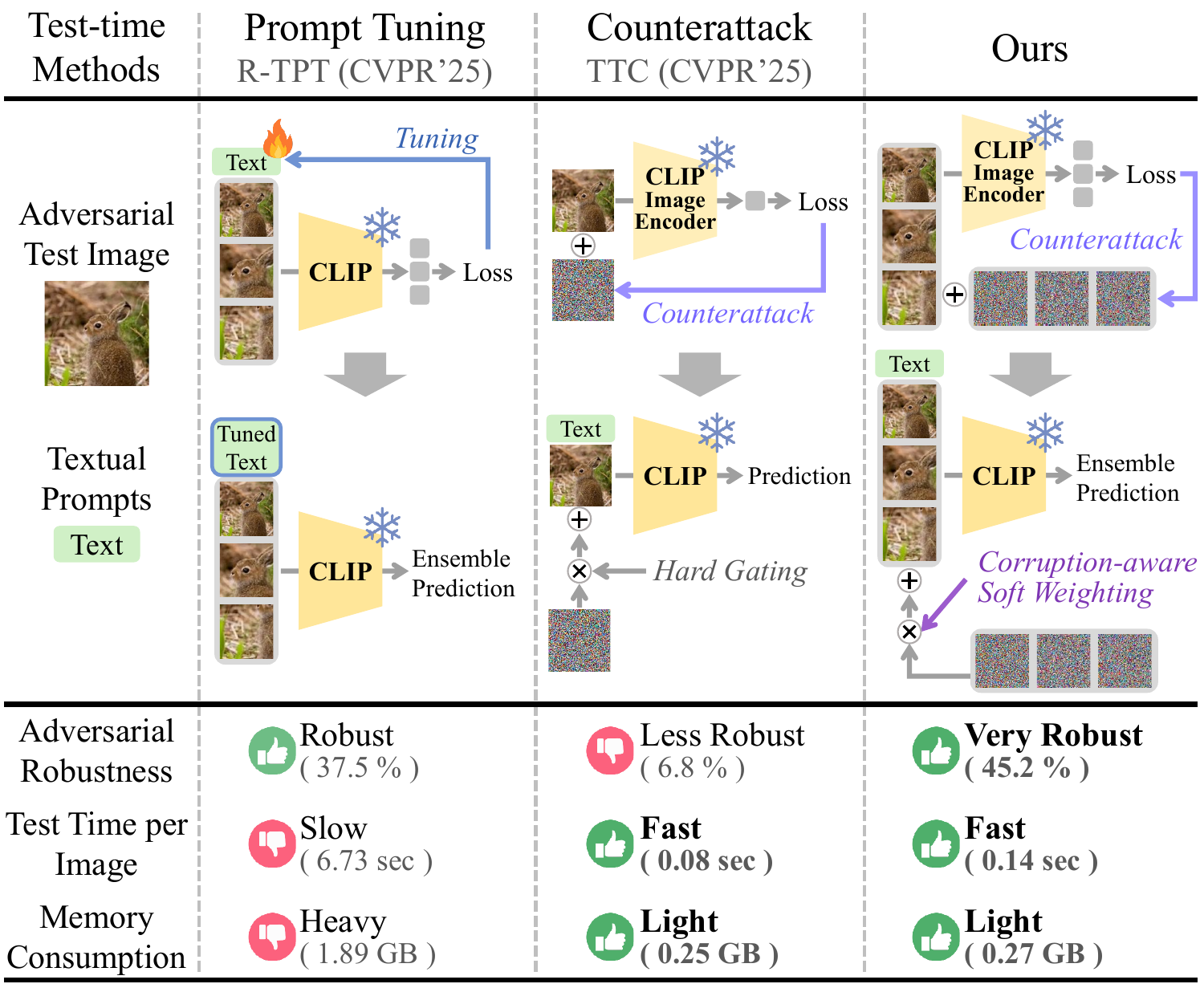}
  \caption{
  Comparison of test-time defense methods for CLIP under strong 100-step PGD attack with a perturbation budget of 4.
  Results are averaged over ten fine-grained recognition datasets.
  Test-time prompt tuning (\ie, R-TPT) incurs high computational and memory overhead, whereas test-time counterattack (TTC) yields limited robustness under strong attacks.
  In contrast, our proposed method achieves substantially higher robustness while maintaining fast inference speed and memory efficiency.
  }
\label{fig:concept-art}
\end{figure}

Vision-language models (VLMs) learn joint image-text representations that enable strong zero-shot generalization across diverse tasks and domains, attracting growing attention from both academia and industry~\cite{zhang2024vision,jia2021scaling,liu2023visual,ramesh2021zero,saharia2022photorealistic,yu2022coca,radford2021learning,kim2024gaussian}.
Among them, CLIP~\cite{radford2021learning} stands out as a representative model that achieves remarkable zero-shot recognition through contrastive pre-training, inspiring numerous extensions and applications~\cite{song2022clip,zhong2022regionclip,yao2023detclipv2}.
Despite its impressive generalization, CLIP remains highly vulnerable to adversarial perturbations, where tiny and invisible changes to an image can cause the model to misclassify the image~\cite{li2024language,li2024one,mao2022understanding,schlarmann2024robust,wang2024pre,zhang2024adversarial,zhou2024few}.
With the increasing deployment of vision-language models in real-world scenarios, ensuring their safety and reliability against adversarial attacks has emerged as a critical challenge, demanding attention from the research community.

\begin{figure}[t!]
  \centering
  \includegraphics[width=0.93\linewidth]{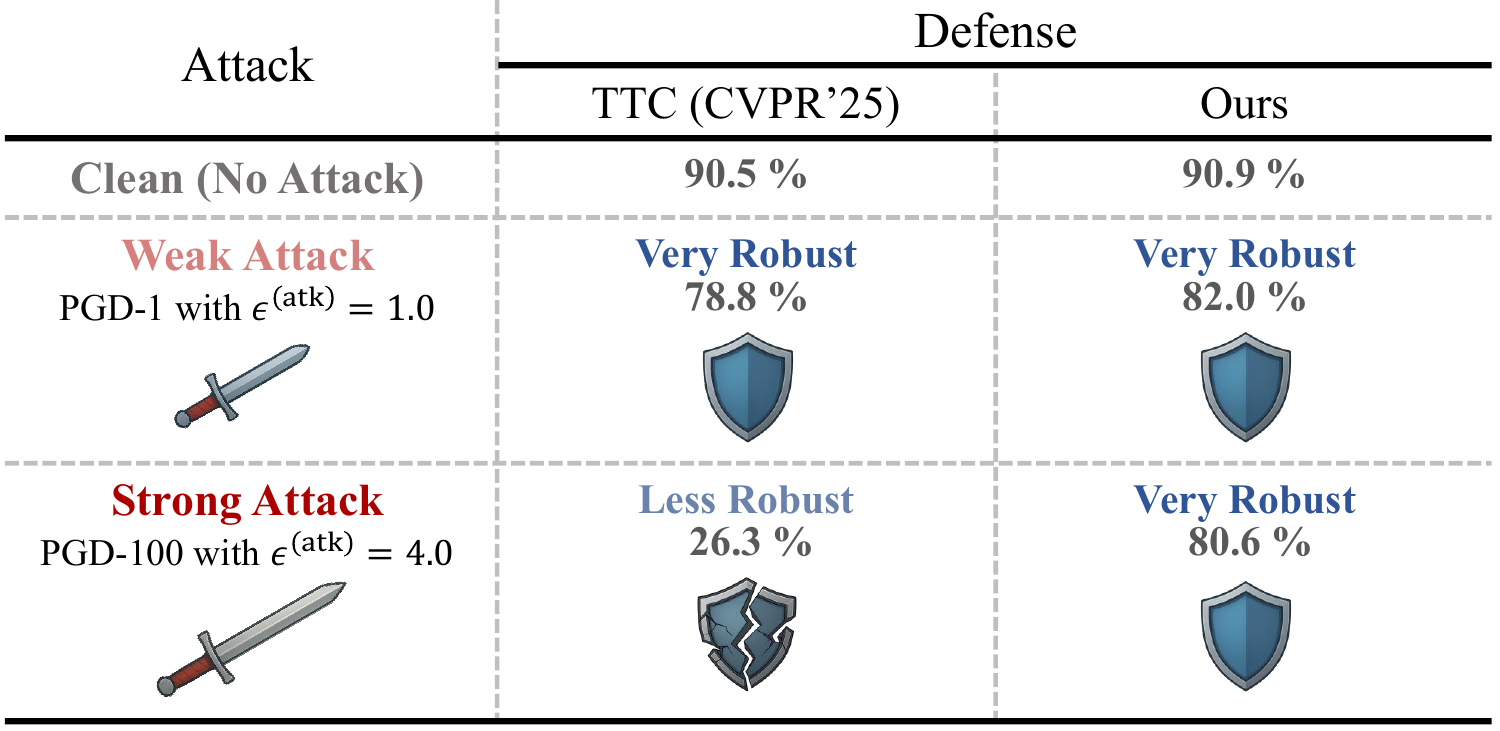}
  \caption{
  Comparison under weak (PGD-1 with a small perturbation budget) and strong (PGD-100 with a large budget) attacks on the Caltech101 dataset.
  TTC exhibits a substantial performance degradation under strong perturbations, whereas our MAC consistently maintains high robustness across both attack strengths.
  }
\label{fig:weak-strong}
\end{figure}

To improve the robustness of CLIP against such perturbations, adversarial fine-tuning approaches have been explored~\cite{mao2022understanding,schlarmann2024robust,wang2024pre,li2024one,zhang2024adversarial}.
While these approaches enhance robustness, they rely on task-specific labeled data, limiting their scalability.
To preserve the zero-shot flexibility of CLIP, recent work has shifted toward test-time defenses, such as test-time prompt tuning (TPT)~\cite{shu2022test,wang2025tapt,sheng2025r} and test-time counterattack (TTC)~\cite{xing2025clip}, which adjust models to adversarial perturbations during inference without requiring labels or retraining.
TPT-based defense methods~\cite{wang2025tapt,sheng2025r} adapt textual prompts by updating prompt parameters during inference, which requires per-instance tuning.
However, this per-instance tuning prevents batch processing of multiple inputs and requires storing gradients and optimizer states, leading to slow inference and high memory consumption.
Unlike these TPT-based methods, TTC~\cite{xing2025clip} achieves efficient inference by counteracting adversarial perturbations using CLIP's pretrained features without any tuning process.
Specifically, TTC performs a counterattack by iteratively perturbing the input to steer its embedding away from a corrupted state.
However, we observe that TTC remains highly vulnerable under strong adversarial attacks, as shown in~\cref{fig:weak-strong}.

We identify two main factors for this vulnerability: (i) \emph{original view guidance}, where the counterattack relies on the embedding of the directly corrupted original image;
and (ii) \emph{noise-driven hard-gating scheme} that decides whether to initiate a counterattack based on noise-induced embedding deviation, which does not reflect the actual corruption severity and hinders adaptive adjustment of the counterattack intensity.
To address these limitations, we propose Multi-view guided Adaptive Counterattack (MAC), a tuning-free test-time defense that performs (i) \emph{multi-view guided counterattacks} and applies (ii) \emph{corruption-aware soft weighting}.
MAC first generates diverse augmented views from an input image and extracts their embeddings using a pretrained CLIP encoder.
MAC then performs multi-view guided counterattacks that jointly steer all views away from their corrupted states, providing reliable guidance by mitigating the reliance on a perturbed original view.
For adaptive counterattack, we define a new measure termed corruption degree, which estimates corruption severity from embedding deviations under stochastic augmentations. Building upon this, MAC applies corruption-aware soft weights that scale the counterattack intensity for each view according to its corruption degree, amplifying the intensity on highly corrupted views while suppressing it on weakly corrupted or clean ones.
Finally, MAC aggregates the counterattacked views through a multi-view ensemble to produce the final prediction.

We comprehensively evaluate MAC against state-of-the-art test-time defenses to assess its robustness and efficiency under a challenging adversarial setting, as summarized in \cref{fig:concept-art}.
MAC achieves substantially stronger robustness than TTC~\cite{xing2025clip}, while being significantly faster and more memory-efficient than the TPT-based defense~\cite{sheng2025r}.
In summary, our main contributions are as follows.
\begin{itemize}
\item We propose \textbf{MAC}, a tuning-free test-time defense for CLIP that performs \emph{multi-view guided counterattacks} to mitigate the reliance on a directly perturbed original view and provide more reliable guidance.
\item We introduce a new measure, \textit{corruption degree}, which effectively captures per-view corruption severity, and based on this measure, we propose \textit{corruption-aware soft weighting} that adaptively controls the counterattack intensity, enabling robust defense even under strong attacks.
\item We demonstrate that MAC is \emph{adversarially robust}, \emph{fast}, and \emph{memory efficient} through extensive evaluations across 20 datasets and diverse attack scenarios.
\end{itemize}
\begin{figure*}[t!]
  \centering
  \includegraphics[width=0.8\linewidth]{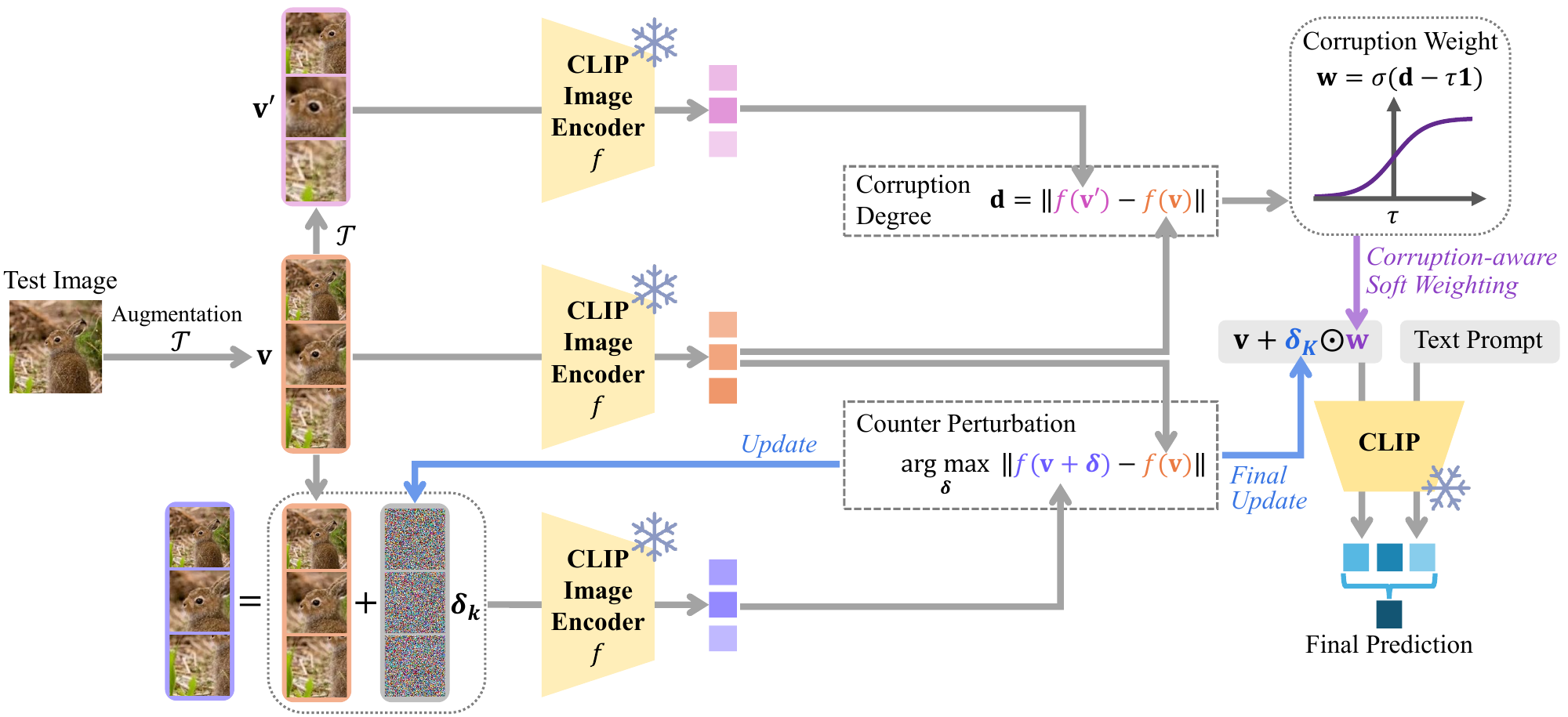}
  \caption{Pipeline of the proposed MAC framework. MAC enhances CLIP's adversarial robustness by leveraging multiple augmented views and adaptive counterattacks.
  Multi-view embeddings are obtained through stochastic augmentations and used to guide view-wise counterattack perturbations.
  Corruption-aware soft weights then adjust the counterattack intensity based on estimated corruption degrees, and the final prediction is produced by aggregating counterattacked views via a multi-view ensemble.
  }
\label{fig:framework}
\end{figure*}

\section{Related Work}
\label{sec:related-work}

\paragraph{Adversarial Robustness.}
Vision models are known to be vulnerable to adversarial examples, where imperceptible perturbations cause mispredictions~\cite{szegedy2013intriguing,goodfellow2014explaining}.
Various adversarial attack strategies have been developed under different threat models \cite{madry2017towards,carlini2017towards,croce2020reliable,moosavi2016deepfool,moosavi2017universal,su2019one,ilyas2018black,andriushchenko2020square,xie2019improving,kurakin2018adversarial,papernot2016limitations,athalye2018synthesizing,athalye2018obfuscated,xie2025chain}.
Among them, Projected Gradient Descent (PGD) \cite{madry2017towards} is widely adopted as a baseline while AutoAttack \cite{croce2020reliable} provides a reliable, parameter-free ensemble of strong attacks.
To enhance adversarial robustness, adversarial training remains prevalent, where adversarial examples are incorporated during training~\cite{bai2021recent,rice2020overfitting,madry2017towards,zhang2019theoretically,wu2020adversarial}. 
Despite its effectiveness, this approach incurs significant training overhead~\cite{shafahi2019adversarial,wong2020fast}.
Beyond adversarial training, other work enhances robustness through training-time augmentation, inference-time purification, and randomized smoothing~\cite{samangouei2018defense,nie2022diffusion,yoon2021adversarial,cohen2019certified,salman2020denoised,salman2019provably,schneider2020improving,wu2021attacking}.
Among them, Wu \etal~\cite{wu2021attacking} demonstrated that perturbing images using the gradient from the cross-entropy loss across all classes can improve robustness.
Despite these advancements, prior efforts have targeted vision-only models. As vision-language models become increasingly deployed in real-world applications~\cite{kim2026gbo,song2022clip,kim2024learnable,kim2022swag}, improving their adversarial robustness has emerged as an urgent and understudied challenge.

\paragraph{Adversarially Robust VLMs.}
Recently, various studies have investigated the adversarial robustness of vision-language models (VLMs)~\cite{shayegani2023jailbreak,zhao2023evaluating,tong2025zero,zhu2025enhancing}. Adversarial fine-tuning, such as contrastive adversarial training~\cite{mao2022understanding,schlarmann2024robust,wang2024pre} and adversarial prompt tuning~\cite{li2024one,zhang2024adversarial}, improves the robustness of VLMs but requires labeled data.
To overcome the dependence on supervision, test-time prompt tuning (TPT)-based defenses~\cite{wang2025tapt,sheng2025r}, inspired by test-time adaptation~\cite{wang2020tent,sun2020test,zhou2025bayesian,huang2025cosmic,karmanov2024efficient,tomar2023tesla,zhang2021tip,wang2022continual,shu2022test,sharifdeen2025tpt,chen2025multi,jeong2025test}, adapt text prompts without labeled data or retraining.
For example, R-TPT~\cite{sheng2025r} tunes prompts by minimizing point-wise entropy for each input during inference.
However, because the prompts are adapted for every single instance, these methods suffer from slow inference when processing large-scale data.
In contrast, tuning-free strategies have also been explored. MeanShift-based augmentation (MTA)~\cite{zanella2024test} aggregates embeddings of multiple views, while the test-time counterattack (TTC)~\cite{xing2025clip} directly leverages CLIP's pretrained features to counteract adversarial perturbations.
Notably, TTC achieves robustness with efficient batch-level inference and no additional tuning, highlighting the potential of tuning-free defenses for VLMs.
However, TTC relies on a single corrupted original view and a hard-gating scheme, which limits its robustness under strong attacks.
To address these issues, our MAC introduces a multi-view guided counterattack paradigm with corruption-aware soft weighting, achieving stronger robustness while preserving the efficiency of tuning-free inference.

\section{Methodology}
\label{sec:method}

Before introducing our proposed framework, we provide the preliminaries necessary to understand our approach, including the formulation of CLIP-based classification, adversarial attack, and test-time counterattack mechanisms.

\subsection{Preliminaries}
\label{sec:prelim}
\paragraph{Classification of CLIP.}
CLIP~\cite{radford2021learning} is a vision-language model that aligns visual and textual modalities in a shared embedding space, allowing zero-shot recognition by matching an image to the most semantically relevant text prompt.
Let $f(\cdot)$ and $g(\cdot)$ denote image and text encoders of CLIP, respectively, and $\phi(\cdot)$ denotes a prompt-template function that maps a class label $c_j$ to a textual prompt $\phi(c_j)$. For an input image $x \in \mathbb{R}^{H \times W \times 3}$, where $H$ and $W$ represent the image height and width, respectively, the similarity score is
\begin{equation}
s_j(x) \;=\; 
\frac{\langle f(x),\, g(\phi(c_j)) \rangle}
{\|f(x)\|_2\,\|g(\phi(c_j))\|_2},
\label{eq:clip-similarity}
\end{equation}
where $\langle \cdot, \cdot \rangle$ denotes the dot product and $\|\cdot\|_2$ denotes the $\ell_2$ norm.
The probability of $x$ belonging to class $c_j$ is computed by normalizing the similarity scores across all classes. The final prediction is then obtained by selecting the class with the highest probability.

\paragraph{Adversarial attack and test-time counterattack.}
Despite strong generalization, CLIP is vulnerable to adversarial perturbations~\cite{goodfellow2014explaining,madry2017towards,szegedy2013intriguing}. When the attacker has full access to CLIP's parameters, a small perturbation $\delta^{(\mathrm{atk})}\in \mathbb{R}^{H \times W \times 3}$ bounded by an $\ell_p$-norm constraint can be optimized to induce misclassification, which is defined as
\begin{equation}
\delta^{(\mathrm{atk})} \;=\;
\mathop{\arg\max}_{\|\delta\|_p \le \epsilon^{(\mathrm{atk})}} 
\; \mathcal{L}(x + \delta,\, y_{\text{gt}}),
\label{eq:adv-attack}
\end{equation}
where $y_{\text{gt}}$ denotes the ground-truth label of $x$, $\mathcal{L}$ is the cross-entropy loss, and $\epsilon^{(\mathrm{atk})}$ represents the attack budget that ensures imperceptibility to human vision.  
PGD~\cite{madry2017towards} solves this optimization iteratively, updating $\delta$.
The resulting adversarial image is obtained by adding the optimized perturbation to the clean image:
\begin{math}
x \;\leftarrow\; x + \delta^{(\mathrm{atk})}.   
\label{eq:adv-image}
\end{math}

To defend CLIP against adversarial perturbations, the test-time counterattack (TTC)~\cite{xing2025clip} leverages CLIP's pre-trained vision encoder without any tuning.
TTC formulates a counterattack perturbation $\delta^{(\mathrm{ttc})}\in \mathbb{R}^{H \times W \times 3}$ that steers the perturbed representations $f(x + \delta)$ away from the potential corrupted state $f(x)$ in the embedding space:
\begin{equation}
\delta^{(\mathrm{ttc})}
= \mathop{\arg\max}_{\|\delta\|_p \le \epsilon^{(\mathrm{ca})}}
\big\| f(x + \delta) - f(x) \big\|_2,
\label{eq:ttc}
\end{equation}
where $\epsilon^{(\mathrm{ca})}$ is the counterattack budget.
TTC iteratively adjusts the input $x$ at test time to avoid corrupted representations, thereby guiding it toward more reliable embeddings.

\subsection{Overview of MAC}
\label{sec:overview}

Although TTC enhances the adversarial robustness of CLIP, we find that it remains highly vulnerable under strong attacks, as shown in~\cref{fig:weak-strong}. This vulnerability mainly stems from two factors. First, TTC relies solely on the perturbed version of the original image, whose embedding becomes unreliable when highly corrupted and thus cannot effectively guide the counterattack.
Second, TTC leverages a noise-driven hard-gating mechanism that measures the change in the image embedding under small random-noise injection to decide whether to initiate a counterattack.
However, such noise-driven deviation is insufficient to assess corruption severity since it cannot reflect structured distortions.
Moreover, this hard-gating mechanism overlooks the potential of corruption severity as a key to adaptively controlling the counterattack intensity.

To this end, we propose Multi-view guided Adaptive Counterattack (MAC), a tuning-free test-time framework for CLIP, as illustrated in \cref{fig:framework}.
Our MAC framework leverages diverse augmented views to obtain reliable representations and introduces a corruption-aware soft weighting mechanism that adaptively scales the counterattack in a view-wise manner.
Specifically, MAC first constructs multiple augmented views of an input image to obtain diverse representations (\cref{sec:multi-view-aug}).
These views are then used to perform a multi-view guided counterattack that refines corrupted embeddings by optimizing view-wise counterattack perturbations (\cref{sec:mv-ttc}).
Next, we define a new measure called \textit{corruption degree}, which captures the structured variations of each view, and, building on this, we perform \textit{corruption-aware soft weighting} that adaptively controls the counterattack intensity (\cref{sec:soft-weight}).
Finally, a multi-view ensemble integrates predictions from adaptively counterattacked views to yield a robust final decision (\cref{sec:mv-ens}).

\subsection{Multi-View Augmentation}
\label{sec:multi-view-aug}
Given an input image $x$, we construct a multi-view $\mathbf{v}\in \mathbb{R}^{N \times H \times W \times 3}$, where $N$ is the total number of views, consisting of the original image $x$ and its $N-1$ augmented variants.
Formally, we define the multi-view as
\begin{math}
\mathbf{v} = \big[\,v_0,\; v_1,\; \ldots,\; v_{N-1}\,\big]^\top,
\label{eq:multi-view}
\end{math}
where $v_0$ is the original image $x$, and $v_i = T_i(x)$ for $i \ge 1$ denotes the $i$-th augmented view obtained by applying an augmentation transformation $T_i$ to $x$.
For the transformation, we sample independent transforms $T_i \sim \mathcal{T}$, where $\mathcal{T}$ includes random affine, color jitter, Gaussian blur, and additive Gaussian noise.
Each view $v_i$ is then encoded into the embedding space through the CLIP image encoder $f(\cdot)$.
We denote the multi-view embedding as
\begin{math}
f(\mathbf{v})
\in \mathbb{R}^{N \times D},
\end{math}
where $D$ is the dimension of the CLIP embedding space.
This multi-view representation captures diverse visual perspectives that enrich feature variations.

\subsection{Multi-View Guided Counterattack}
\label{sec:mv-ttc}
Given the multi-view $\mathbf{v}$ and its corresponding embeddings $f(\mathbf{v})$, our objective is to estimate multi-view guided counterattack perturbations $\bm{\delta}^{(\mathrm{mvc})}\in \mathbb{R}^{N \times H \times W \times 3}$
that steer the counterattack-perturbed representations away from their corrupted states in the embedding space across $N$ views:
\begin{equation}
\begin{aligned}
\bm{\delta}^{(\mathrm{mvc})}
&=
\arg\max_{\bm{\delta}}
\big\|
f(\mathbf{v} + \bm{\delta}) - f(\mathbf{v})
\big\|_F, \\
\text{s.t.}\quad
&\|\delta_i\|_p \le \epsilon^{(\mathrm{ca})},
\quad \forall i,
\end{aligned}
\label{eq:mvc}
\end{equation}
where $\|\cdot\|_F$ denotes the Frobenius norm, $\delta_i\in \mathbb{R}^{H \times W \times 3}$ denotes the $i$-th element of $\bm{\delta}$, and $\epsilon^{(\mathrm{ca})}$ is the counterattack budget.
The optimization is performed for $K$ iterations using projected gradient updates, 
which iteratively refines the perturbations so that each view is guided away from its corrupted representation and toward a more reliable embedding.
At iteration $k$, we perform a gradient ascent step on the objective for each view and then project onto the $\ell_p$ ball, updating perturbation $\bm{\delta}_k^{(\mathrm{mvc})}$.
We initialize $\bm{\delta}_0^{(\mathrm{mvc})}$ by sampling each element uniformly from $[-\epsilon^{(\mathrm{ca})},\epsilon^{(\mathrm{ca})}]$.
After $K$ iterations, the final multi-view guided counterattack perturbations $\bm{\delta}_K^{(\mathrm{mvc})}$ are obtained, 
which are subsequently applied to produce the counterattacked multi-view inputs.

\subsection{Corruption-aware Soft Weighting}
\label{sec:soft-weight}
In the multi-view guided counterattack, we should consider that images and their views are not equally corrupted in practice for two reasons: 
(i) \emph{Per-view heterogeneity.} The attacker perturbs the original image directly, and the augmented views derived from the original do not necessarily inherit the same corruption strength as the original; some become only partially or weakly corrupted due to the transformations.
(ii) \emph{Unknown test-time conditions.} In real deployments, input images can be clean, weakly attacked, or strongly attacked, but the model does not know this corruption level.
Consequently, applying the same counterattack perturbation intensity to all images and their views can lead to over-correction 
for weakly corrupted or clean ones and under-correction 
for strongly corrupted ones. 
To address this, we introduce a \emph{corruption-aware soft weighting} mechanism that adaptively scales the counterattack perturbation according to the estimated corruption degree of each view.

\paragraph{Multi-view embeddings.}
To estimate the corruption severity of each view $v_i$, we measure the sensitivity of its embedding
to augmentations drawn from the augmentation distribution $\mathcal{T}$.
First, we generate an augmented multi-view $\mathbf{v}'\in \mathbb{R}^{N \times H \times W \times 3}$ as
\begin{math}
\mathbf{v}' = \big[\,T_0'(v_0),\; T_1'(v_1),\; \ldots,\; T_{N-1}'(v_{N-1})\,\big]^\top,
\label{eq:aug-multi-view}
\end{math}
where $T_i'\sim \mathcal{T}$ is an augmentation transformation applied to $v_i$.
We obtain the augmented multi-view embeddings $f(\mathbf{v}')\in \mathbb{R}^{N \times D}$ through the CLIP image encoder $f(\cdot)$.

\begin{figure}[t!]
  \centering
  \includegraphics[width=0.9\linewidth]{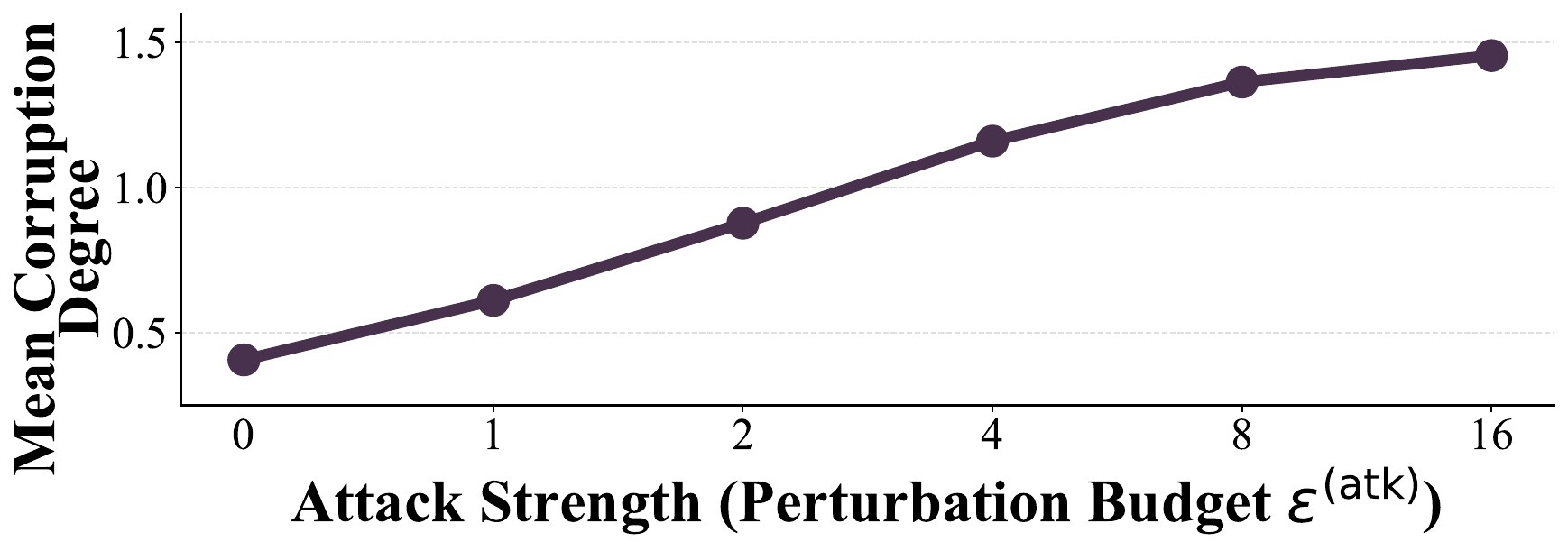}
  \caption{
  Mean corruption degree measured on the ImageNet dataset under varying attack strengths.
  The perturbation budget $\epsilon^{(\mathrm{atk})}=0$ corresponds to clean inputs, and larger $\epsilon^{(\mathrm{atk})}$ values indicate stronger adversarial attacks.
  The corruption degree increases monotonically with attack strength, showing that it effectively reflects the severity of corruption in images.
  }
\label{fig:corruption-degree}
\end{figure}

\paragraph{Corruption degree.} Subsequently, we define a new measure called \textit{corruption degree}, which can serve as an indicator of the corruption severity of each view.
\begin{equation}
d_i =  \big\|\,\tilde f_i(T_i'(v_i)) - \tilde f_i(v_i)\,\big\|_2,
\label{eq:corruption-dir}
\end{equation}
where $\tilde f_i(\cdot) = \frac{f(\cdot)}{\|f(v_i)\|_2}$ denotes the embedding normalized by the $\ell_2$ norm of the $i$-th view, allowing the measure to capture augmentation sensitivity relative to the original view's embedding scale.
As shown in \cref{fig:corruption-degree}, the mean corruption degree increases monotonically as the attack strength grows. Clean or weakly attacked images produce only small embedding deviations under stochastic augmentations, whereas strongly attacked images yield much larger deviations.
Hence, the corruption degree $d_i$ can serve as an indicator of the corruption severity of the $i$-th view.

\paragraph{Corruption-aware soft weighting.} We then map $\mathbf{d}=[d_0,\,\ldots\,,d_{N-1}]^\top$ into a corruption-aware weight $\mathbf{w}\in[0,1]^N$ via a sigmoid activation:
\begin{equation}
\mathbf{w} = \sigma\!\left(\frac{\mathbf{d} - \tau_{\text{thres}}}{\tau_{\text{temp}}}\right),
\label{eq:weight}
\end{equation}
where $\sigma(\cdot)$ denotes the element-wise sigmoid function, 
$\tau_{\text{thres}}$ is the corruption threshold, and $\tau_{\text{temp}}$ is a temperature parameter.
This soft weighting enables smooth transitions between clean and corrupted inputs.
Finally, the corruption-aware weight $\mathbf{w}$ modulates the counterattack perturbations $\bm{\delta}_K^{(\mathrm{mvc})}$, and the final counterattacked multi-view input $\mathbf{v}^{(\mathrm{mvc})}\in \mathbb{R}^{N \times H \times W \times 3}$ is reconstructed as
\begin{equation}
\mathbf{v}^{(\mathrm{mvc})} 
= \mathbf{v} + \bm{\delta}_K^{(\mathrm{mvc})} \odot \mathbf{w},
\label{eq:mv-final}
\end{equation}
where $\odot$ denotes the element-wise product, and $\mathbf{w}\in\mathbb{R}^{N}$ is broadcast along spatial and channel dimensions.
This operation amplifies the counterattack on highly corrupted views 
while suppressing it on weakly corrupted or clean views.
As a result, we adaptively adjust the counterattack intensity, guiding each view toward its own reliable representation.

\subsection{Multi-View Ensemble}
\label{sec:mv-ens}
Given the counterattacked multi-view input $\mathbf{v}^{(\mathrm{mvc})} = [\,v_0^{(\mathrm{mvc})}, \ldots, v_{N-1}^{(\mathrm{mvc})}\,]^\top$,
we aggregate per-view CLIP predictions into a single one to integrate complementary cues across views for a stable and robust prediction.
For each class $c_j$, we compute the similarity score $s_j(v_i^{(\mathrm{mvc})})$ of the $i$-th view using~\cref{eq:clip-similarity}.
We then average the per-view similarity scores and obtain the final prediction $\hat{y}$ via a softmax:
\begin{equation}
\bar{{s}}_j
=
\frac{1}{N}\sum_{i=0}^{N-1}{s}_j\big(v_i^{(\mathrm{mvc})}\big), \;\; 
\hat{y} = \arg\max_{j}\frac{\exp(\bar{s}_j)}{\sum_l\exp(\bar{s}_l)}.
\label{eq:mve-ensemble}
\end{equation}
\begin{table*}[t]
\caption{
Evaluation of various test-time adaptation methods on ten fine-grained recognition datasets using CLIP-ViT-B/32, reporting both clean accuracy (Acc.) and adversarial accuracy under PGD-100 attack with $\epsilon$ = 4.0 (Rob.). The highest score is highlighted in bold. $\Delta$ Rob. represents the robust accuracy gain of our method over the best existing tuning-free method.
}
\centering
\small
\resizebox{\textwidth}{!}{
\setlength{\tabcolsep}{6.5pt}
\begin{tabular}{lll|cccccccccc | c}
\toprule
Category & Method & Metric & Caltech101 & DTD & Flower102 & Pets & UCF101 & Aircraft & EuroSAT & Cars & SUN397 & Food101 & Average \\
\midrule
\multirow{2}{*}{Baseline} 
  & \multirow{2}{*}{CLIP~\cite{radford2021learning}} 
  & Acc. & 91.3 & 42.1 & 63.2 & 85.3 & 61.6 & 18.2 & 29.5 & 58.8 & 61.6 & 77.3 & 58.9 \\
  &  & \cellcolor{robgreen}Rob. 
        & \cellcolor{robgreen}0.1 
        & \cellcolor{robgreen}0.0 
        & \cellcolor{robgreen}0.0 
        & \cellcolor{robgreen}0.0 
        & \cellcolor{robgreen}0.0 
        & \cellcolor{robgreen}0.0 
        & \cellcolor{robgreen}0.0 
        & \cellcolor{robgreen}0.0 
        & \cellcolor{robgreen}0.0 
        & \cellcolor{robgreen}0.0 
        & \cellcolor{robgreen}0.0 \\
\midrule
\multirow{4}{*}{Tuning-based}
  & \multirow{2}{*}{TAPT~\cite{wang2025tapt}}  
  & Acc. & 90.4 & 39.2 & 60.3 & 82.5 & \textbf{64.4} & 16.3 & \textbf{42.5} & \textbf{62.5} & 60.0 & \textbf{85.8} & \textbf{60.4} \\
  &  & \cellcolor{robgreen}Rob. & \cellcolor{robgreen}38.7 & \cellcolor{robgreen}12.4 & \cellcolor{robgreen}9.0 & \cellcolor{robgreen}14.6 & \cellcolor{robgreen}6.4 & \cellcolor{robgreen}2.6 & \cellcolor{robgreen}7.8& \cellcolor{robgreen}7.4&  \cellcolor{robgreen}18.4& \cellcolor{robgreen}5.8& 12.3 \cellcolor{robgreen}  \\ \cmidrule{2-14}
  & \multirow{2}{*}{R\mbox{-}TPT~\cite{sheng2025r}} 
  & Acc. & 90.6 & 42.1 & 62.6 & 84.5 & 62.8 & 19.1 & 28.2 & 60.9 & 63.7 & 78.3 & 59.3 \\
  &  & \cellcolor{robgreen}Rob. 
        & \cellcolor{robgreen}78.0 
        & \cellcolor{robgreen}29.3 
        & \cellcolor{robgreen}39.0 
        & \cellcolor{robgreen}56.5 
        & \cellcolor{robgreen}42.4 
        & \cellcolor{robgreen}9.5 
        & \cellcolor{robgreen}5.2 
        & \cellcolor{robgreen}28.0 
        & \cellcolor{robgreen}43.9 
        & \cellcolor{robgreen}43.1 
        & \cellcolor{robgreen}37.5 \\
\midrule
\multirow{8}{*}{Tuning-free} 
  & \multirow{2}{*}{MTA~\cite{zanella2024test}} 
  & Acc. & \textbf{91.4} & \textbf{43.2} & \textbf{63.5} & \textbf{85.9} & 62.8 & \textbf{20.3} & 30.3 & 61.0 & \textbf{63.8} & 78.8 & 60.1 \\
  &  & \cellcolor{robgreen}Rob. 
        & \cellcolor{robgreen}69.1 
        & \cellcolor{robgreen}15.7 
        & \cellcolor{robgreen}22.9 
        & \cellcolor{robgreen}47.8 
        & \cellcolor{robgreen}32.8 
        & \cellcolor{robgreen}2.4 
        & \cellcolor{robgreen}0.4 
        & \cellcolor{robgreen}10.5 
        & \cellcolor{robgreen}26.9 
        & \cellcolor{robgreen}29.7 
        & \cellcolor{robgreen}25.8 \\
        \cmidrule{2-14}
  & \multirow{2}{*}{TTC~\cite{xing2025clip}} 
  & Acc. & 90.5 & 38.9 & 61.1 & 81.9 & 62.0 & 15.3 & 30.8 & 52.8 & 56.6 & 76.0 & 56.6 \\
  &  & \cellcolor{robgreen}Rob. & \cellcolor{robgreen}26.3 & \cellcolor{robgreen}6.2 & \cellcolor{robgreen}4.0 & \cellcolor{robgreen}11.2 & \cellcolor{robgreen}6.9 & \cellcolor{robgreen}0.7 & \cellcolor{robgreen}2.5 & \cellcolor{robgreen}2.3 & \cellcolor{robgreen}4.6 & \cellcolor{robgreen}3.6 & \cellcolor{robgreen}6.8 \\ \cmidrule{2-14}
  & \multirow{2}{*}{MAC (Ours)} 
  & Acc. & 90.9 & 41.7 & 62.8 & 85.2 & 62.3 & 18.2 & 29.4 & 58.4 & 61.2 & 76.6 & 58.7 \\
  &  & \cellcolor{robgreen}Rob. 
        & \cellcolor{robgreen}\textbf{80.6} 
        & \cellcolor{robgreen}\textbf{32.9} 
        & \cellcolor{robgreen}\textbf{45.6} 
        & \cellcolor{robgreen}\textbf{68.0} 
        & \cellcolor{robgreen}\textbf{48.2} 
        & \cellcolor{robgreen}\textbf{15.5} 
        & \cellcolor{robgreen}\textbf{11.4} 
        & \cellcolor{robgreen}\textbf{38.2}     
        & \cellcolor{robgreen}\textbf{51.7} 
        & \cellcolor{robgreen}\textbf{60.3} 
        & \cellcolor{robgreen}\textbf{45.2} \\\cmidrule{2-14}
  & \multicolumn{2}{l|}{\cellcolor{robgreen}$\Delta$ Rob.} 
    & \cellcolor{robgreen}\textcolor{teal}{\textbf{+11.5}}
    & \cellcolor{robgreen}\textcolor{teal}{\textbf{+17.2}}
    & \cellcolor{robgreen}\textcolor{teal}{\textbf{+22.7}}
    & \cellcolor{robgreen}\textcolor{teal}{\textbf{+20.2}}
    & \cellcolor{robgreen}\textcolor{teal}{\textbf{+15.4}}
    & \cellcolor{robgreen}\textcolor{teal}{\textbf{+13.1}}
    & \cellcolor{robgreen}\textcolor{teal}{\textbf{+8.9}}
    & \cellcolor{robgreen}\textcolor{teal}{\textbf{+27.7}}
    & \cellcolor{robgreen}\textcolor{teal}{\textbf{+24.8}}
    & \cellcolor{robgreen}\textcolor{teal}{\textbf{+30.6}}
    & \cellcolor{robgreen}\textcolor{teal}{\textbf{+19.2}} \\
\bottomrule
\end{tabular}}
\label{tab:vit_finegrained_sota_comparisons}
\end{table*}

\begin{table*}[t]
\caption{
Evaluation of various test-time adaptation methods on ImageNet and four ImageNet OOD benchmarks using CLIP-ViT-B/32, reporting both clean accuracy (Acc.) and adversarial accuracy under PGD-100 attack with $\epsilon$ = 4.0 (Rob.). The highest score is highlighted in bold. $\Delta$ Rob. represents the robust accuracy gain of our method over the best existing tuning-free method.
}
\centering
\small
\resizebox{0.79\textwidth}{!}{
\setlength{\tabcolsep}{6.5pt}
\begin{tabular}{lll|ccccc|c}
\toprule
Category & Method & Metric & ImageNet & ImageNet-A & ImageNet-V2 & ImageNet-R & ImageNet-S & Average \\
\midrule
\multirow{2}{*}{Baseline} 
  & \multirow{2}{*}{CLIP~\cite{radford2021learning}} 
  & Acc. & 62.2 & 30.4 & 55.0 & 65.4 & 39.8 & 50.6 \\
  &  & \cellcolor{robgreen}Rob. & \cellcolor{robgreen}0.0 & \cellcolor{robgreen}0.0 & \cellcolor{robgreen}0.0 & \cellcolor{robgreen}0.0 & \cellcolor{robgreen}0.0 & \cellcolor{robgreen}0.0 \\
\midrule
\multirow{2}{*}{Tuning-based} 
  & \multirow{2}{*}{R\mbox{-}TPT~\cite{sheng2025r}} 
  & Acc. & 64.8 & 37.1 & 57.8 & 69.0 & 40.4 & 53.8 \\
  &  & \cellcolor{robgreen}Rob. & \cellcolor{robgreen}37.1 & \cellcolor{robgreen}11.7 & \cellcolor{robgreen}31.0 & \cellcolor{robgreen}49.8 & \cellcolor{robgreen}22.7 & \cellcolor{robgreen}30.5 \\
\midrule
\multirow{8}{*}{Tuning-free} 
  & \multirow{2}{*}{MTA~\cite{zanella2024test}} 
  & Acc. & \textbf{65.7} & \textbf{37.9} & \textbf{58.1} & \textbf{69.2} & \textbf{41.8} & \textbf{54.5} \\
  &  & \cellcolor{robgreen}Rob. & \cellcolor{robgreen}24.2 & \cellcolor{robgreen}5.5 & \cellcolor{robgreen}18.7 & \cellcolor{robgreen}32.8 & \cellcolor{robgreen}9.4 & \cellcolor{robgreen}18.1 \\ \cmidrule{2-9}
  & \multirow{2}{*}{TTC~\cite{xing2025clip}} 
  & Acc. & 51.8 & 30.0 & 49.2 & 61.3 & 35.3 & 45.5 \\
  &  & \cellcolor{robgreen}Rob. & \cellcolor{robgreen}4.7 & \cellcolor{robgreen}0.6 & \cellcolor{robgreen}4.1 & \cellcolor{robgreen}8.4 & \cellcolor{robgreen}5.4 & \cellcolor{robgreen}4.6 \\ \cmidrule{2-9}
  & \multirow{2}{*}{MAC (Ours)} 
  & Acc. & 61.7 & 29.1 & 54.8 & 65.5 & 39.6 & 50.1 \\
  &  & \cellcolor{robgreen}Rob. & \cellcolor{robgreen}\textbf{47.6} & \cellcolor{robgreen}\textbf{14.8} & \cellcolor{robgreen}\textbf{40.4} & \cellcolor{robgreen}\textbf{55.2} & \cellcolor{robgreen}\textbf{33.5} & \cellcolor{robgreen}\textbf{38.3} \\ \cmidrule{2-9}
  & \multicolumn{2}{l|}{\cellcolor{robgreen}$\Delta$ Rob.} 
  & \cellcolor{robgreen}\textcolor{teal}{\textbf{+23.4}} 
  & \cellcolor{robgreen}\textcolor{teal}{\textbf{+9.3}} 
  & \cellcolor{robgreen}\textcolor{teal}{\textbf{+21.7}} 
  & \cellcolor{robgreen}\textcolor{teal}{\textbf{+22.4}} 
  & \cellcolor{robgreen}\textcolor{teal}{\textbf{+24.1}} 
  & \cellcolor{robgreen}\textcolor{teal}{\textbf{+20.2}} \\
\bottomrule
\end{tabular}}
\label{tab:vit_imagenet_sota_comparisons}
\end{table*}

\section{Experiment}
\label{sec:experiment}

\subsection{Experimental Setup}

\paragraph{Datasets and Evaluation Metrics.}
For evaluation, we use a wide range of fine-grained classification benchmarks: Caltech101~\cite{fei2004learning}, DTD~\cite{cimpoi2014describing}, Flower102~\cite{nilsback2008automated}, Pets~\cite{parkhi2012cats}, UCF101~\cite{soomro2012dataset}, Aircraft~\cite{maji2013fine}, EuroSAT~\cite{helber2019eurosat},
Cars~\cite{krause20133d},
SUN397~\cite{xiao2010sun}, and Food101~\cite{bossard2014food}. 
For large-scale and out-of-distribution (OOD) evaluation, we additionally use ImageNet~\cite{deng2009imagenet} along with four ImageNet-based distribution-shift benchmarks:
ImageNet-A~\cite{hendrycks2021natural}, ImageNet-V2~\cite{recht2019imagenet}, ImageNet-R~\cite{hendrycks2021many}, and ImageNet-S~\cite{wang2019learning}.
These fine-grained and ImageNet datasets span diverse domains such as objects, animals, plants, materials, remote sensing imagery, human activities, and scene understanding.
In supplementary material, we further conducted experiments on five additional datasets, including CIFAR10~\cite{krizhevsky2009learning}, CIFAR100~\cite{krizhevsky2009learning}, STL10~\cite{coates2011analysis}, Caltech256~\cite{griffin2007caltech}, and Country211~\cite{radford2021learning}. 

We evaluate performance using two standard metrics adopted in prior works~\cite{xing2025clip,sheng2025r}:
Acc., the average classification accuracy on clean test samples, and
Rob., the average accuracy under adversarial perturbations, which quantifies robustness against attacks.
Since our method operates under a test-time paradigm, all evaluations are performed exclusively on the test splits without using any training data.

\paragraph{Compared Methods}
As a baseline, we use the zero-shot inference results of CLIP~\cite{radford2021learning}.
We compare MAC against state-of-the-art methods in adversarially robust test-time adaptation for CLIP, including both tuning-based methods~\cite{wang2025tapt,sheng2025r} and tuning-free methods~\cite{zanella2024test,xing2025clip}.
For fair comparisons, all methods are evaluated under the same strong attack scenario, using their officially released implementations and default hyperparameters.

\begin{figure*}[t!]
  \centering
  \includegraphics[width=0.85\linewidth]{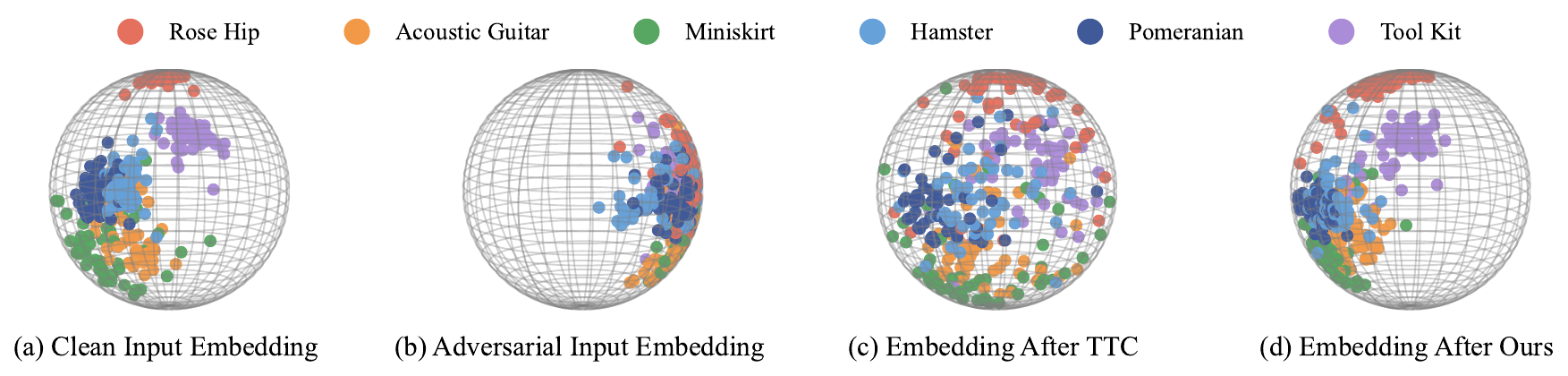}
  \caption{Visualization of clean, adversarial, and counterattacked embeddings from six ImageNet classes. 
  Compared to TTC, our MAC restores class-level separation even under strong adversarial perturbations, yielding a feature distribution similar to that of clean embeddings.}
\label{fig:visual_feats}
\vspace{-0.2em}
\end{figure*}

\paragraph{Implementation Details}
For MAC, we use pretrained CLIP ViT-B/32 as the vision-language backbone, and all CLIP parameters are frozen without any tuning process.
For each test image, we generate $N=2$ views.
Inspired by AugMix~\cite{hendrycks2019augmix}, the augmentation distribution $\mathcal{T}$ applies a mild random affine transformation to every image, followed by additional augmentations, including Gaussian blur, additive Gaussian noise, and color jitter, each independently applied in a stochastic manner.
Robustness is measured under an $\ell_\infty$ PGD attack~\cite{madry2017towards} crafted directly on the frozen CLIP model.
For strong adversarial attack settings, we set the number of PGD steps to $100$, step size to $1$, and $\epsilon^{(\mathrm{atk})}$ to $4$.
Since the attacker's settings are unknown to the defense algorithm, MAC employs counterattack settings independent of the attacker.
Although a large $\epsilon^{(\mathrm{ca})}$ does not affect the external input as MAC's perturbations operate entirely inside the defense pipeline, we simply set a reasonable $\ell_\infty$ budget of $\epsilon^{(\mathrm{ca})}=8$ with short iterations ($K{=}4$), which preserves image quality.
For corruption-aware soft weighting, we use a sigmoid with threshold $\tau_{\text{thres}}{=}0.7$ and temperature $\tau_{\text{temp}}{=}0.01$.
All experiments are implemented in PyTorch and run using a batch size of 128 on a single A100 GPU.
More details are included in the supplementary material.

\begin{table}[t]
\caption{Comparison of average adversarial robustness of test-time adaptation methods under diverse attack scenarios, including DI\textsuperscript{2}-FGSM~\cite{xie2019improving}, CW~\cite{carlini2017towards}, AutoAttack~\cite{croce2020reliable}, and MAC-adaptive attacks, on ten fine-grained recognition datasets.}
\centering
\small
\resizebox{\linewidth}{!}{
\begin{tabular}{ll|cccc}
\toprule
Category & Method & DI\textsuperscript{2} & CW & AutoAttack & MAC-adaptive\\
\midrule
\multirow{1}{*}{Baseline}
  & CLIP~\cite{radford2021learning} & 0.0 & 0.1 & 0.1 & 0.0 \\
\midrule
\multirow{1}{*}{Tuning\mbox{-}based}
  & R\mbox{-}TPT~\cite{sheng2025r} & 20.8 & 41.4 & 43.4 & \textbf{36.2} \\
\midrule
\multirow{3}{*}{Tuning\mbox{-}free}
  & MTA~\cite{zanella2024test} & 14.5 & 31.7 & 32.5 & 29.8 \\
  & TTC~\cite{xing2025clip} & 5.1 & 42.3 & 8.4 & 2.4 \\
  & MAC (Ours) & \textbf{21.3} & \textbf{47.5} & \textbf{47.4} & 17.5 \\
\bottomrule
\end{tabular}}
\label{tab:diverse_attacks}
\vspace{-0.3em}
\end{table}


\subsection{Comparisons with State-of-the-Art}
\label{sec:comparisons}

\paragraph{Adversarial Robustness.}
As shown in \cref{tab:vit_finegrained_sota_comparisons}, our MAC attains the highest adversarial robustness with an average robust accuracy of \textbf{45.2\%} across 10 fine-grained recognition datasets, outperforming all competing methods. 
Notably, despite being a \emph{tuning-free} approach, MAC surpasses the strongest tuning-based baseline, R-TPT, by \textbf{+7.7} points on average.
While clean accuracy remains competitive with prior work, MAC consistently improves robustness across all datasets by a large margin, yielding gains of up to \textbf{+30.6} points over the best existing tuning-free method. These results confirm that MAC delivers substantial robustness improvement without sacrificing clean accuracy.
On ImageNet and its OOD variants, a similar trend is observed as shown in \cref{tab:vit_imagenet_sota_comparisons}, where MAC again achieves the best robustness, improving over the strongest tuning-free baseline by \textbf{+20.2} points on average.
This demonstrates MAC's strong effectiveness under OOD settings.

\paragraph{Robustness under diverse attacks.}
We further evaluate MAC against various adversarial attack scenarios, which include DI\textsuperscript{2}-FGSM~\cite{xie2019improving}, CW~\cite{carlini2017towards}, AutoAttack~\cite{croce2020reliable}, and adaptive white-box attacks designed to differentiate through our MAC pipeline using BPDA~\cite{athalye2018obfuscated,bengio2013estimating}, as shown in \cref{tab:diverse_attacks}.
MAC consistently achieves the highest robustness across DI\textsuperscript{2}-FGSM, CW, and AutoAttack. 
Notably, MAC demonstrates stronger robustness than TTC even under the MAC-adaptive attack, where the adversary is fully aware of the MAC pipeline and optimizes against it directly. 
These results highlight MAC's superior resilience to both standard and adaptive adversarial threats, demonstrating its robustness beyond the conventional PGD setting.
More detailed attack settings are provided in the supplementary material.

\begin{table}[t]
\caption{
Comparison of average memory usage, inference speed, and adversarial robustness of test-time adaptation methods on ten fine-grained recognition datasets using CLIP-ViT-B/32.  
}
\centering
\small
\resizebox{\columnwidth}{!}{
\begin{tabular}{ll|ccc}
\toprule
Category & Method & Memory (GB) & Speed (s/img) & Rob. (\%) \\
\midrule
\multirow{1}{*}{Baseline}
  & CLIP~\cite{radford2021learning} & \textbf{0.18} & \textbf{0.06} & 0.0 \\
\midrule
\multirow{2}{*}{Tuning-based}
  & TAPT~\cite{wang2025tapt} & 8.87 & 3.57 & 12.3 \\
  & R\mbox{-}TPT~\cite{sheng2025r} & 1.89 & 6.73 & 37.5 \\
\midrule
\multirow{3}{*}{Tuning-free}
  & MTA~\cite{zanella2024test} & 0.23 & 6.68 & 25.8 \\
  & TTC~\cite{xing2025clip} & 0.25 & 0.08 & 6.8 \\
  & MAC (Ours) & 0.27 & 0.14 & \textbf{45.2} \\
\bottomrule
\end{tabular}}
\label{tab:efficiency}
\vspace{-0.2em}
\end{table}

\begin{figure*}[t!]
  \centering
  \begin{minipage}[t]{0.32\linewidth}
    \centering
    \includegraphics[width=\linewidth]{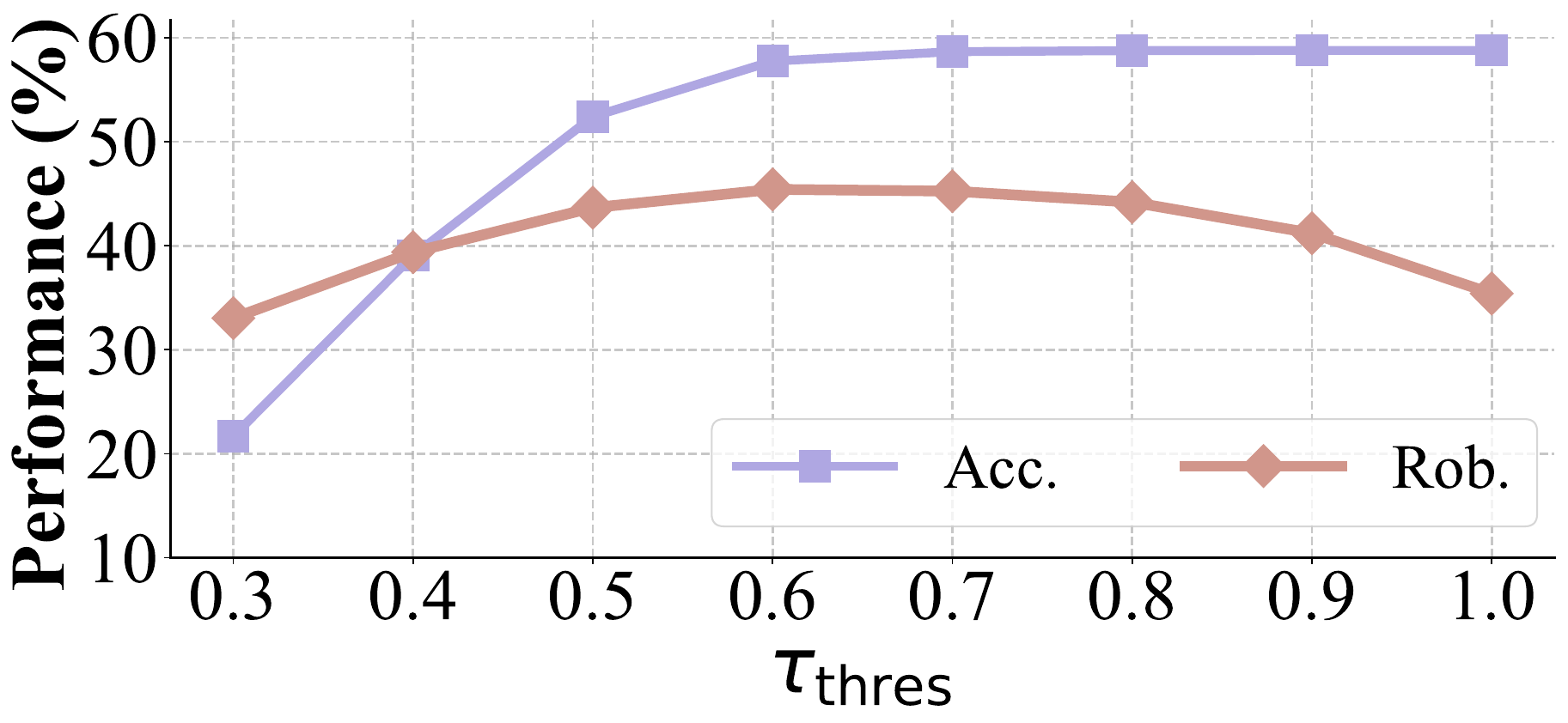}
    \vspace{-2em}
    \caption*{(a)}
  \end{minipage}
  \hfill
  \begin{minipage}[t]{0.32\linewidth}
    \centering
    \includegraphics[width=\linewidth]{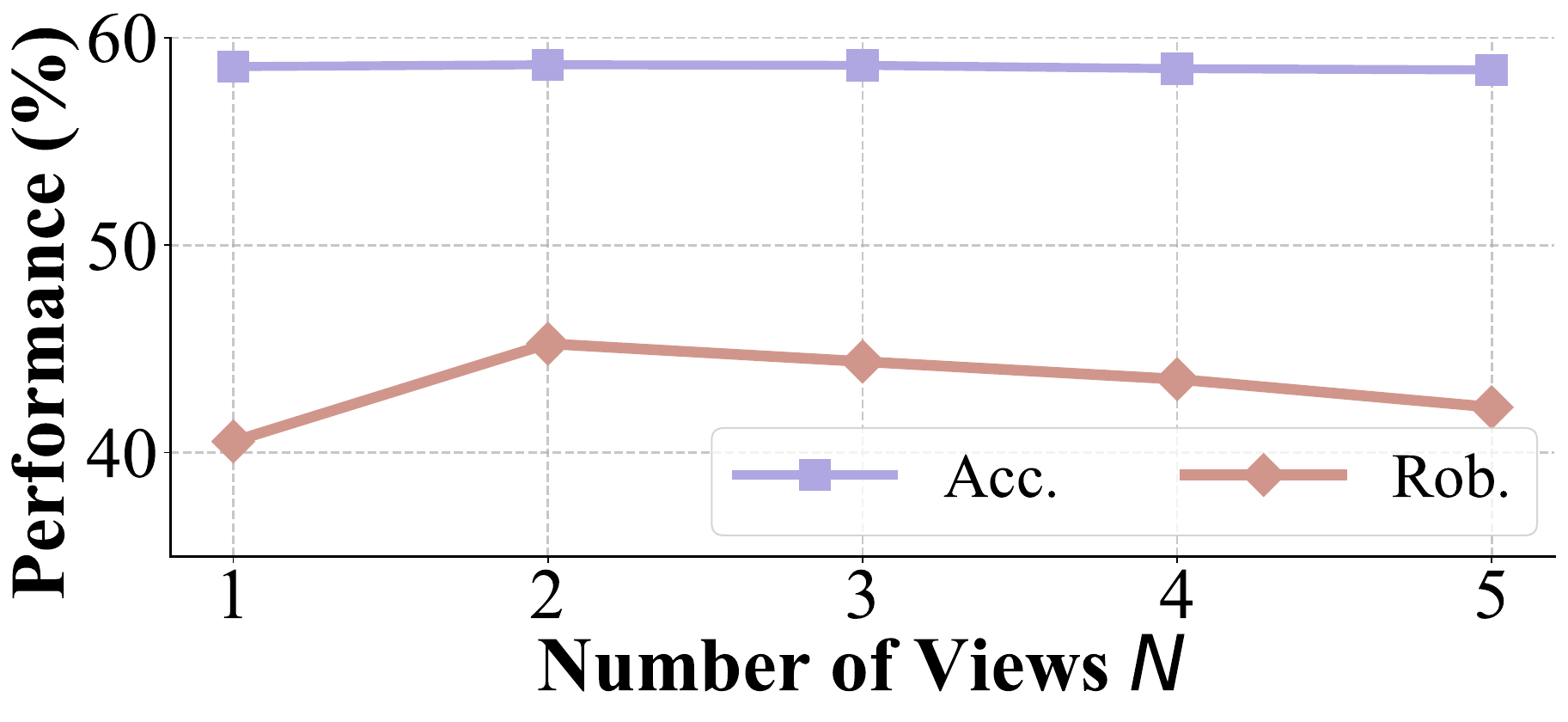}
    \vspace{-2em}
    \caption*{(b)}
  \end{minipage}
  \hfill
  \begin{minipage}[t]{0.32\linewidth}
    \centering
    \includegraphics[width=\linewidth]{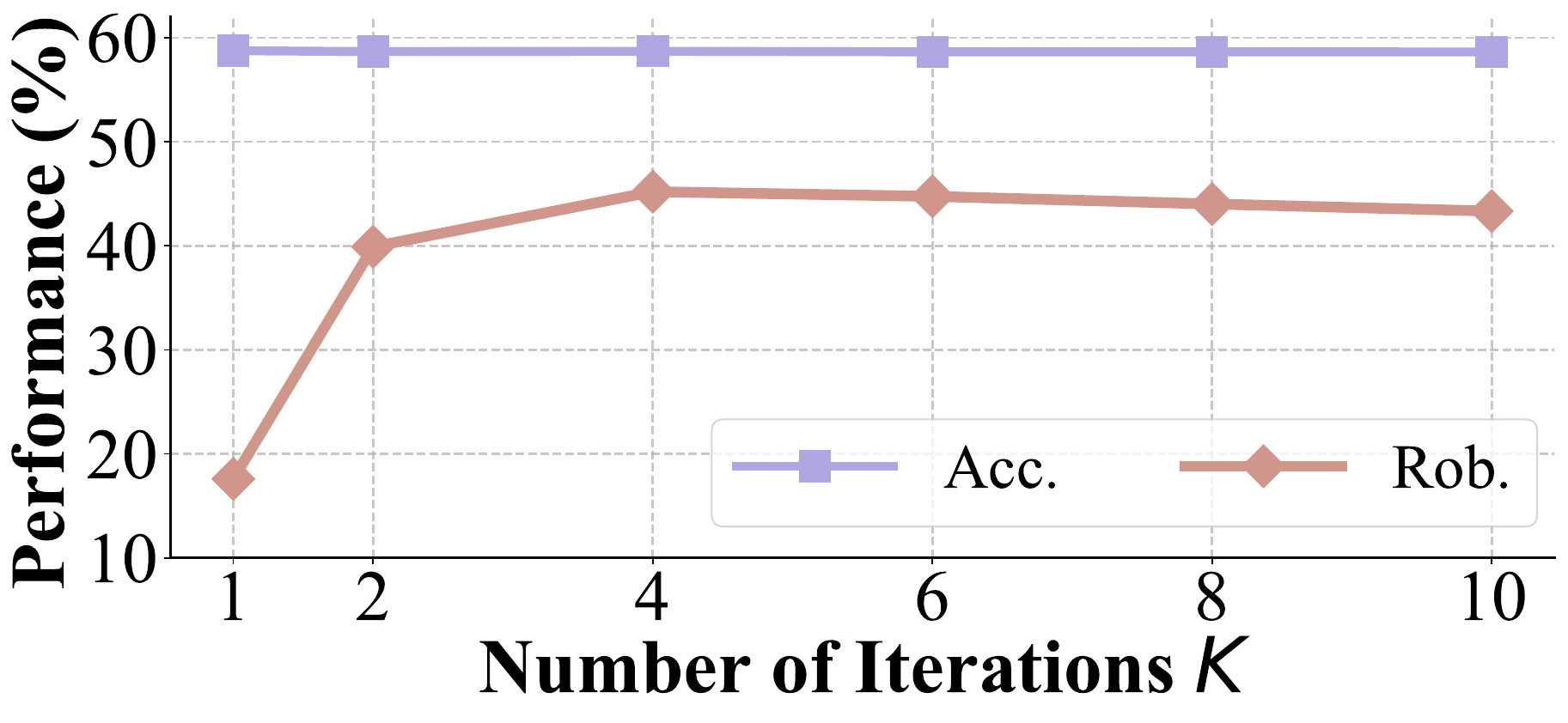}
    \vspace{-2em}
    \caption*{(c)}
  \end{minipage}
  \vspace{-0.7em}
  \caption{
  Analysis with respect to key hyperparameters: (a) threshold $\tau_{\text{thres}}$ for corruption-aware soft weighting, (b) the number of views $N$, and (c) the number of counterattack iterations $K$. Performance is averaged over ten fine-grained recognition datasets.}
  \label{fig:ablation_core}
  \vspace{-0.2em}
\end{figure*}
 
\begin{figure}[t!]
  \centering
  \includegraphics[width=0.9\linewidth]{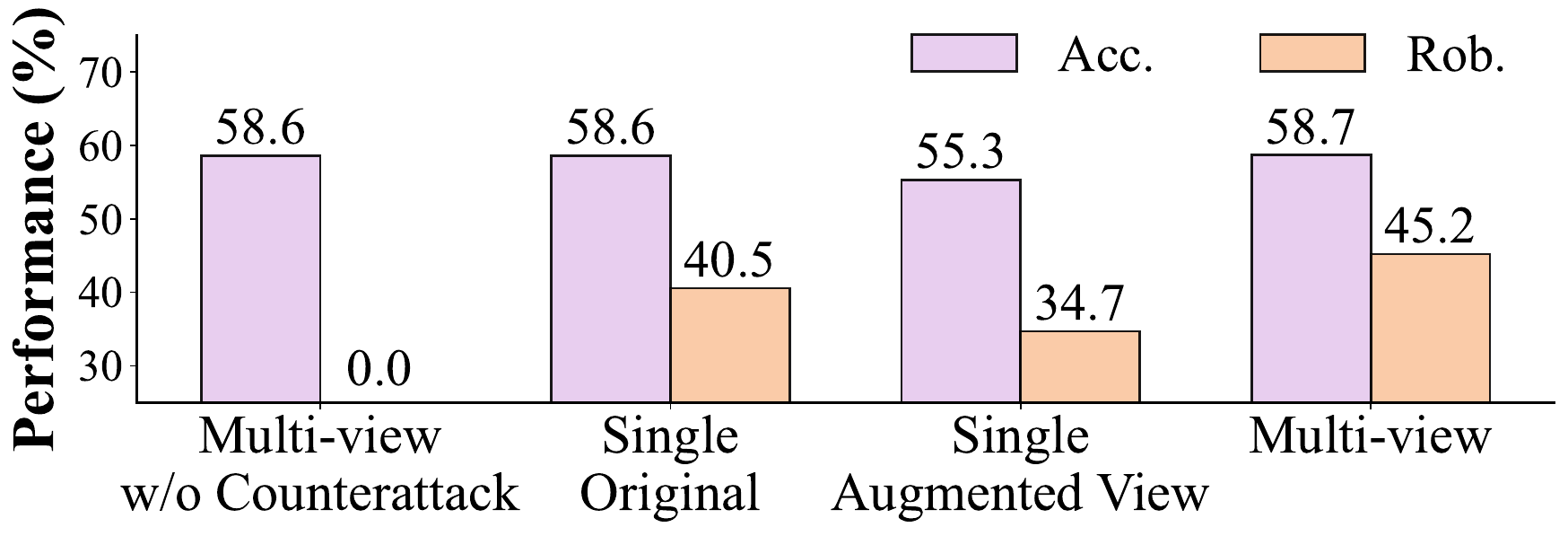}
  \vspace{-0.5em}
  \caption{Ablation study of the multi-view guided counterattack, averaged over ten fine-grained recognition datasets.}
\label{fig:ablation_aug}
\vspace{-0.2em}
\end{figure}

\paragraph{Efficiency and memory overhead.}
As summarized in \cref{tab:efficiency}, we report the memory usage and inference speed of test-time adaptation methods. 
We measure GPU memory overhead per input image as the difference between peak and initial allocation during inference. 
Tuning-based methods require storing gradients and optimizer states, and inherently operate at the instance level, which prevents batch processing of multiple inputs.
This leads to high memory consumption and slow inference.
Among tuning-free methods, MTA operates at the instance level, as it requires a large number of augmented views (up to 127) per input, leading to slow processing despite its moderate memory usage. 
In contrast, the tuning-free counterattack design of TTC and MAC enables efficient batch processing, adapting multiple inputs simultaneously.
Consequently, MAC achieves low memory usage (\textbf{0.27\,GB}) and fast inference (\textbf{0.14\,s/img}) while delivering the highest robustness among all compared methods, demonstrating its superior efficiency.

\subsection{In-depth Analysis}
\label{sec:analysis}

\paragraph{Feature-space visualization.}
\cref{fig:visual_feats} visualizes the per-sample embeddings of clean, adversarial, and counterattacked images of TTC and MAC using t-SNE~\cite{maaten2008visualizing}. 
Under strong PGD-100 attack with $\epsilon^{(\mathrm{atk})}=4$, the feature distribution becomes severely distorted and class boundaries collapse. 
After counterattack, TTC still produces a feature distribution with poor class separation, while MAC recovers a well-separated distribution that closely resembles the clean embedding manifold. This shows MAC's ability to restore semantically coherent representations that remain discriminative for classification, even under strong attacks.

\paragraph{Effect of multi-view.}
\cref{fig:ablation_aug} demonstrates that \emph{multi-view guided counterattack} delivers clear robustness gains while preserving clean accuracy, indicating that multi-view guidance is effective. Single-view baselines (original or a single augmented view) achieve reasonable performance, but when multiple views are combined, their synergy yields substantially stronger robustness. In contrast, simply aggregating multiple views \emph{without} counterattack offers little benefit and remains brittle under adversarial perturbations.

\begin{figure}[t!]
  \centering
  \includegraphics[width=0.9\linewidth]{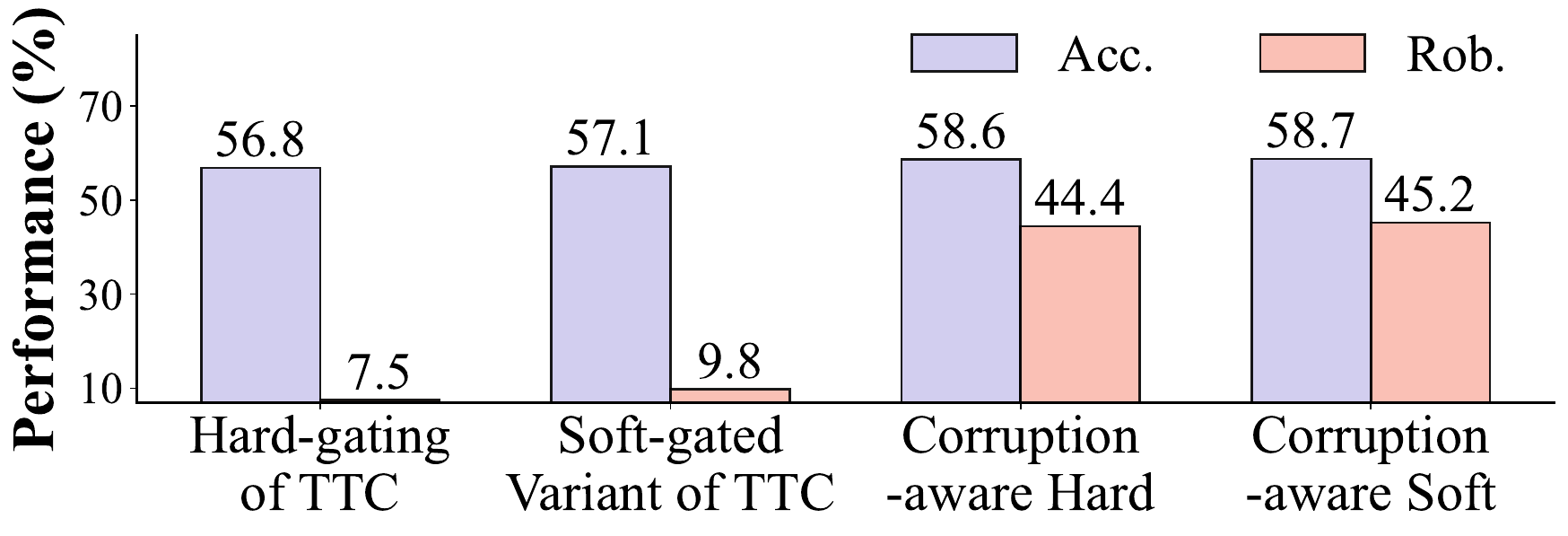}
  \vspace{-0.5em}
  \caption{Ablation study of the corruption-aware soft weighting, averaged over ten fine-grained recognition datasets.}
  \vspace{-0.4em}
\label{fig:ablation_masking}
\end{figure}

\paragraph{Effect of corruption-aware soft weighting.}
\cref{fig:ablation_masking} compares different masking schemes in MAC. Hard-gating of TTC~\cite{xing2025clip} decides whether to initiate a counterattack based on a noise-driven deviation. Both the TTC hard rule and its soft-gated variant yield very low robustness because such noise-driven deviation is insufficient to assess corruption severity, as it fails to capture structured distortions.
In contrast, our corruption-aware mechanism achieves higher robustness, as the \emph{corruption degree} captures structured variations, providing a reliable measure of corruption severity.
Among them, the \emph{corruption-aware soft weighting} is most effective, adaptively adjusting the counterattack intensity based on the corruption degree of each view.

\paragraph{Hyperparameters.}
\cref{fig:ablation_core} shows how the performance of MAC varies with three key hyperparameters: (a) threshold $\tau_{\text{thres}}$ for corruption weights, (b) the number of views $N$, and (c) the number of counterattack iterations $K$.
\textbf{(a) Threshold $\tau_{\text{thres}}$:} When $\tau_{\text{thres}}$ is low, counterattacks are applied even to clean images, which degrades clean accuracy. As $\tau_{\text{thres}}$ increases, clean performance gradually improves and saturates around $\tau_{\text{thres}}{=}0.7$. 
Robustness peaks at $\tau_{\text{thres}}{=}0.7$. A low $\tau_{\text{thres}}$ triggers strong counterattacks even for weakly perturbed samples, whereas a high $\tau_{\text{thres}}$ weakens the counterattack for strongly corrupted ones, slightly reducing robustness in both cases. Nevertheless, average robustness remains state-of-the-art (above 30\%) across the entire $\tau_{\text{thres}}$ range, indicating that MAC is less sensitive to the choice of $\tau_{\text{thres}}$.
\textbf{(b) Views $N$:} Moving from a single view to multiple views improves robustness by nearly 5 percentage points, confirming the benefit of multi-view guidance for counterattacks.
The gain is small beyond two views as excessive views introduce redundant or conflicting information. Considering both accuracy and efficiency, using two views offers the best trade-off for MAC.
\textbf{(c) Iterations $K$:} When $K{=}1$, the counterattack perturbation is insufficiently optimized, resulting in low robustness. As $K$ increases, robustness improves and saturates beyond $K{=}4$, indicating that only a few counterattack iterations are sufficient for MAC, whereas the PGD attack uses 100 iterations.

\section{Conclusion}
\label{sec:conclusion}
In this work, we present MAC, a tuning-free test-time defense for CLIP that generates diverse augmented views and applies view-wise counterattacks to steer their embeddings away from corrupted states.
For adaptive counterattacks, we introduce a corruption degree that estimates the corruption severity of each view, and, based on this degree, we adjust the counterattack intensity of each view through soft weighting.
The adaptively counterattacked views are then aggregated to produce a robust final prediction. 
Across diverse datasets, OOD benchmarks, and attack conditions, MAC consistently achieves state-of-the-art robustness without sacrificing clean performance. 
Furthermore, its tuning-free design enables highly efficient inference with low latency and memory usage, making MAC readily deployable in real-world systems without retraining or tuning.
We believe that MAC can provide a strong and efficient foundation for future advances in robust and practical VLMs.

\section*{Acknowledgements}
This work was supported by the IITP (Institute of Information \& Communications Technology Planning \& Evaluation)-ITRC (Information Technology Research Center) grant funded by the Korea government (Ministry of Science and ICT) (IITP-2026-2024-00437102, Contribution Rate: 50\%) and the National Research Foundation of Korea (NRF) grant funded by the Korea government (MSIT) (No. RS-2026-25495369, Contribution Rate: 50\%).
\clearpage
\setcounter{page}{1}
\maketitlesupplementary

\begin{table*}[t]
\caption{
Evaluation of various adversarial finetuning and test-time adaptation methods on five datasets using CLIP-ViT-B/32, reporting both clean accuracy (Acc.) and adversarial accuracy under PGD-10 attack with $\epsilon$ = 4.0 (Rob.). The highest score is highlighted in bold.
In the adversarial fine-tuning methods, the superscripts represent the attack budgets used to generate adversarial images during adversarial fine-tuning.
}
\centering
\small
\resizebox{\linewidth}{!}{
\setlength{\tabcolsep}{3pt}
\begin{tabular}{ll|c|ccccccc|cccccc}
\toprule
\multirow{2}{*}{Dataset} & \multirow{2}{*}{Metric} & \multirow{2}{*}{CLIP} &
\multicolumn{7}{c|}{Adversarial Fine-tuning} & \multicolumn{6}{c}{Test-time Defense} \\
\cmidrule(lr){4-10}\cmidrule(lr){11-16}
& & & CLIP-FT & TeCoA$^{1}$ & TeCoA$^{4}$ & PMG-AFT$^{1}$ & PMG-AFT$^{4}$ & FARE$^{1}$ & FARE$^{4}$ & RN & TTE & Anti-adv & HD & TTC & Ours \\
\midrule
\multirow{2}{*}{CIFAR10}
& Acc. & \textbf{85.12} & 84.90 & 64.61 & 65.15 & 70.69 & 71.45 & 74.44 & 78.46 & 81.18 & 84.74 & 83.44 & 78.23 & 81.18 & 83.84 \\
& \cellcolor{robgreen}Rob. 
& \cellcolor{robgreen}0.43 & \cellcolor{robgreen}2.75 & \cellcolor{robgreen}7.69 & \cellcolor{robgreen}11.7 & \cellcolor{robgreen}10.20 & \cellcolor{robgreen}15.59 & \cellcolor{robgreen}1.94 & \cellcolor{robgreen}5.42 & \cellcolor{robgreen}0.00 & \cellcolor{robgreen}3.47 & \cellcolor{robgreen}0.32 & \cellcolor{robgreen}1.67 & \cellcolor{robgreen}28.51 & \cellcolor{robgreen}\textbf{35.96} \\
\multirow{2}{*}{CIFAR100}
& Acc. & 57.14 & \textbf{59.51} & 35.96 & 36.30 & 40.32 & 41.51 & 46.67 & 47.38 & 56.34 & 58.61 & 53.96 & 52.86 & 56.34 & 56.70 \\
& \cellcolor{robgreen}Rob. 
& \cellcolor{robgreen}0.05 & \cellcolor{robgreen}0.67 & \cellcolor{robgreen}6.54 & \cellcolor{robgreen}9.25 & \cellcolor{robgreen}7.60 & \cellcolor{robgreen}10.80 & \cellcolor{robgreen}2.64 & \cellcolor{robgreen}4.54 & \cellcolor{robgreen}0.00 & \cellcolor{robgreen}1.37 & \cellcolor{robgreen}0.22 & \cellcolor{robgreen}0.00 & \cellcolor{robgreen}9.06 & \cellcolor{robgreen}\textbf{18.36} \\
\multirow{2}{*}{STL10}
& Acc. & \textbf{96.40} & 94.49 & 87.40 & 81.69 & 88.56 & 84.35 & 91.72 & 89.11 & 95.85 & 96.26 & 95.47 & 89.50 & 95.83 & 96.17 \\
& \cellcolor{robgreen}Rob. 
& \cellcolor{robgreen}0.16 & \cellcolor{robgreen}3.75 & \cellcolor{robgreen}24.80 & \cellcolor{robgreen}31.83 & \cellcolor{robgreen}28.49 & \cellcolor{robgreen}35.40 & \cellcolor{robgreen}9.99 & \cellcolor{robgreen}17.59 & \cellcolor{robgreen}0.06 & \cellcolor{robgreen}32.56 & \cellcolor{robgreen}2.25 & \cellcolor{robgreen}3.39 & \cellcolor{robgreen}52.40 & \cellcolor{robgreen}\textbf{73.49} \\
\multirow{2}{*}{Caltech256}
& Acc. & 81.72 & 78.53 & 61.14 & 52.05 & 62.24 & 53.32 & 73.32 & 67.22 & 81.25 & \textbf{82.48} & 79.40 & 79.12 & 76.59 & 81.57 \\
& \cellcolor{robgreen}Rob. 
& \cellcolor{robgreen}0.12 & \cellcolor{robgreen}1.41 & \cellcolor{robgreen}8.29 & \cellcolor{robgreen}11.76 & \cellcolor{robgreen}10.65 & \cellcolor{robgreen}13.68 & \cellcolor{robgreen}2.18 & \cellcolor{robgreen}5.09 & \cellcolor{robgreen}0.16 & \cellcolor{robgreen}23.23 & \cellcolor{robgreen}1.44 & \cellcolor{robgreen}0.34 & \cellcolor{robgreen}27.25 & \cellcolor{robgreen}\textbf{52.56} \\
\multirow{2}{*}{Country211}
& Acc. & \textbf{15.25} & 12.07 & 4.75 & 3.66 & 4.64 & 3.34 & 9.26 & 6.58 & 14.80 & 14.66 & 11.60 & 11.72 & 11.99 & 14.45 \\
& \cellcolor{robgreen}Rob. 
& \cellcolor{robgreen}0.00 & \cellcolor{robgreen}0.00 & \cellcolor{robgreen}0.05 & \cellcolor{robgreen}0.19 & \cellcolor{robgreen}0.12 & \cellcolor{robgreen}0.24 & \cellcolor{robgreen}0.00 & \cellcolor{robgreen}0.02 & \cellcolor{robgreen}0.00 & \cellcolor{robgreen}0.24 & \cellcolor{robgreen}0.00 & \cellcolor{robgreen}0.00 & \cellcolor{robgreen}2.44 & \cellcolor{robgreen}\textbf{3.78} \\
\midrule
\multirow{2}{*}{Average}
& Acc. & 67.13 & 65.90 & 50.77 & 47.77 & 53.29 & 50.79 & 59.08 & 57.75 & 65.88 & \textbf{67.35} & 64.77 & 62.29 & 64.39 & 66.55 \\
& \cellcolor{robgreen}Rob. 
& \cellcolor{robgreen}0.15 & \cellcolor{robgreen}1.72 & \cellcolor{robgreen}9.47 & \cellcolor{robgreen}12.95 & \cellcolor{robgreen}11.41 & \cellcolor{robgreen}15.14 & \cellcolor{robgreen}3.35 & \cellcolor{robgreen}6.53 & \cellcolor{robgreen}0.04 & \cellcolor{robgreen}12.17 & \cellcolor{robgreen}0.85 & \cellcolor{robgreen}1.08 & \cellcolor{robgreen}23.93 & \cellcolor{robgreen}\textbf{36.83} \\
\bottomrule
\end{tabular}}
\label{tab:adv_finetuning_comparisons}
\end{table*}

\section{Limitations}
\label{sec:limitation}
Although MAC demonstrates consistent robustness improvements across 20 diverse datasets without any retraining, it is inherently dependent on the choice of the augmentation distribution and the multi-view configuration. In our experiments, we adopt a set of low-level image transformations and a small number of views, which may not be optimal for all visual domains or attack types. Consequently, the quality and diversity of the generated views can bound the achievable robustness, particularly in specialized modalities (e.g., medical or remote-sensing imagery) where appropriate semantics-preserving transformations differ from those of natural images. Exploring richer, domain-aware, or even learnable augmentation strategies that can automatically adapt the multi-view generation process to dataset- or task-specific characteristics is an important direction for future work.

\section{Datasets}
\label{suppl-sec:datasets}

As our method operates in a test-time setting, all experiments are performed strictly on the test splits without using training data.
We evaluate our method on \textbf{20 datasets}:
\begin{itemize}
    \item \textbf{Caltech101}~\cite{fei2004learning}: Object category dataset with 101 classes and 2,465 test images.
    \item \textbf{DTD}~\cite{cimpoi2014describing}: Texture classification dataset with 47 classes and 1,692 test images.
    \item \textbf{Flower102}~\cite{nilsback2008automated}: Flower species dataset with 102 classes and 2,463 test images.
    \item \textbf{Pets}~\cite{parkhi2012cats}: Pet image dataset with 37 classes and 3,669 test images.
    \item \textbf{UCF101}~\cite{soomro2012dataset}: Human action frame-based dataset with 101 classes and 3,783 test images.
    \item \textbf{Aircraft}~\cite{maji2013fine}: Fine-grained aircraft recognition dataset with 100 classes and 3,333 test images.
    \item \textbf{EuroSAT}~\cite{helber2019eurosat}: Satellite imagery dataset with 10 classes and 8,100 test images.
    \item \textbf{Cars}~\cite{krause20133d}: Fine-grained car model classification dataset with 196 classes and 8,041 test images.
    \item \textbf{SUN397}~\cite{xiao2010sun}: Scene recognition dataset with 397 classes and 19,850 test images.
    \item \textbf{Food101}~\cite{bossard2014food}: Food image dataset containing 101 classes and 30,300 test images.
    \item \textbf{ImageNet}~\cite{deng2009imagenet}: Large-scale object classification dataset with 1,000 classes and 50,000 test images.
    \item \textbf{ImageNet-A}~\cite{hendrycks2021natural}: Natural adversarial images with 200 classes and 7,500 test images.
    \item \textbf{ImageNet-V2}~\cite{recht2019imagenet}: Re-collected ImageNet samples with 1,000 classes and 10,000 test images.
    \item \textbf{ImageNet-R}~\cite{hendrycks2021many}: Artistic renditions of ImageNet classes with 200 classes and 30,000 test samples.
    \item \textbf{ImageNet-S}~\cite{wang2019learning}: Sketch images with 1,000 classes and 50,889 test images.

    \item \textbf{CIFAR10}~\cite{krizhevsky2009learning}: 10-class natural image dataset with 10,000 test images.
    \item \textbf{CIFAR100}~\cite{krizhevsky2009learning}: 100-class natural image dataset with 10,000 test images.
    \item \textbf{STL10}~\cite{coates2011analysis}: Image classification dataset with 10 classes and 8,000 test images.
    \item \textbf{Caltech256}~\cite{griffin2007caltech}: Object recognition dataset with 256 classes and 30,607 test images.
    \item \textbf{Country211}~\cite{radford2021learning}: Geographic location prediction dataset with 211 classes and 21,100 test images.
\end{itemize}

\section{Implementation Details}
\label{suppl-sec:details}
\paragraph{Backbones and Baselines.}
We use pretrained CLIP ResNet50, CLIP ViT-B/32, CLIP ViT-B/16, and CLIP ViT-L/14 as the vision-language backbones, and all CLIP parameters are frozen without any tuning process.
The CLIP text prompt is formulated as ``a photo of [CLASS]''.
We use CLIP ViT-B/32 as the default setting for all experiments, unless mentioned otherwise.
For fair comparisons, all compared methods are evaluated under the same CLIP model and strong attack scenario, using their officially released implementations and default hyperparameters.\footnote{MTA: \url{https://github.com/MaxZanella/MTA};\; TTC: \url{https://github.com/Sxing2/CLIP-Test-time-Counterattacks};\; TAPT: \url{https://github.com/xinwong/TAPT};\; R-TPT: \url{https://github.com/TomSheng21/R-TPT}}
Among the compared methods, we conduct TAPT~\cite{wang2025tapt} experiments with CLIP ViT-B/16, as the method provides the necessary precomputed values exclusively for this backbone.

\paragraph{Augmentation and counterattack.}
Inspired by AugMix~\cite{hendrycks2019augmix}, the augmentation distribution $\mathcal{T}$ applies a mild random affine transformation (rotation $\leq 5^\circ$, translation $\leq 4\%$, scale jitter $\pm 8\%$) to every image, followed by additional augmentations, including Gaussian blur with standard deviation 0.5-1.2, additive Gaussian noise with standard deviation $\leq 0.02$, and color jitter (brightness, contrast, and saturation $\pm 0.08$; hue $\pm 0.02$), each applied independently with a probability of $0.5$.
For counterattack, we set a reasonable $\ell_\infty$ budget of $\epsilon^{(\mathrm{ca})}=8$ with short iterations ($K{=}4$) and step size of $2$.
Using this setting, we empirically confirm that the counterattack perturbations remain visually imperceptible, as illustrated in \cref{fig:visual-quality}.

\paragraph{Algorithm}
\cref{alg:mac} provides an overview of our proposed MAC method, detailing how multi-view guided counterattacks and corruption-aware soft weighting are incorporated during test-time inference.

\begin{algorithm}[t]
\caption{MAC: Multi-View Guided Adaptive Counterattack}
\label{alg:mac}
\begin{algorithmic}[1]
\Require CLIP image encoder $f$, text encoder $g$, class prompts $\{\phi(c_j)\}$, test image $x$
\Ensure Robust prediction $\hat{y}$
\State Construct a multi-view $v = [v_0,\dots,v_{N-1}]$ by applying $N-1$ stochastic augmentations to $x$.
\State Obtain view embeddings $z_i = f(v_i)$ for all $i$.
\State Initialize perturbations $\delta_i \sim \mathcal{U}[-\epsilon^{(\mathrm{ca})}, \epsilon^{(\mathrm{ca})}]$ and update $\delta_i$ through $K$ steps of projected gradient ascent on $\|f(v_i+\delta_i)-z_i\|_2^2$ with step size $\alpha$ and $\ell_\infty$-projection $\|\delta_i\|_\infty \le \epsilon^{(\mathrm{ca})}$, yielding $\delta_i$ for each view.
\State Sample an augmentation $T$, compute normalized embeddings $\bar{z}_i = z_i/\|z_i\|_2$ and $\bar{z}_i' = f(T(v_i))/\|z_i\|_2$, set corruption degree $d_i = \|\bar{z}_i' - \bar{z}_i\|_2$, and obtain a soft weight $w_i = \sigma\!\big((d_i - \tau_{\mathrm{thres}})/\tau_{\mathrm{temp}}\big) \in [0,1]$ for all $i$.
\State Form adaptively counterattacked views $\tilde{v}_i = v_i + w_i \delta_i$ for all $i$.
\State Compute CLIP similarities between $f(\tilde{v}_i)$ and $\{g(\phi(c_j))\}$, average scores across all views, and set $\hat{y} = \arg\max_j \bar{s}_j$ where $\bar{s}_j$ is the aggregated similarity for class $j$.
\end{algorithmic}
\end{algorithm}

\begin{figure*}[t!]
  \centering

  \begin{minipage}[t]{0.24\linewidth}
    \centering
    \includegraphics[width=\linewidth]{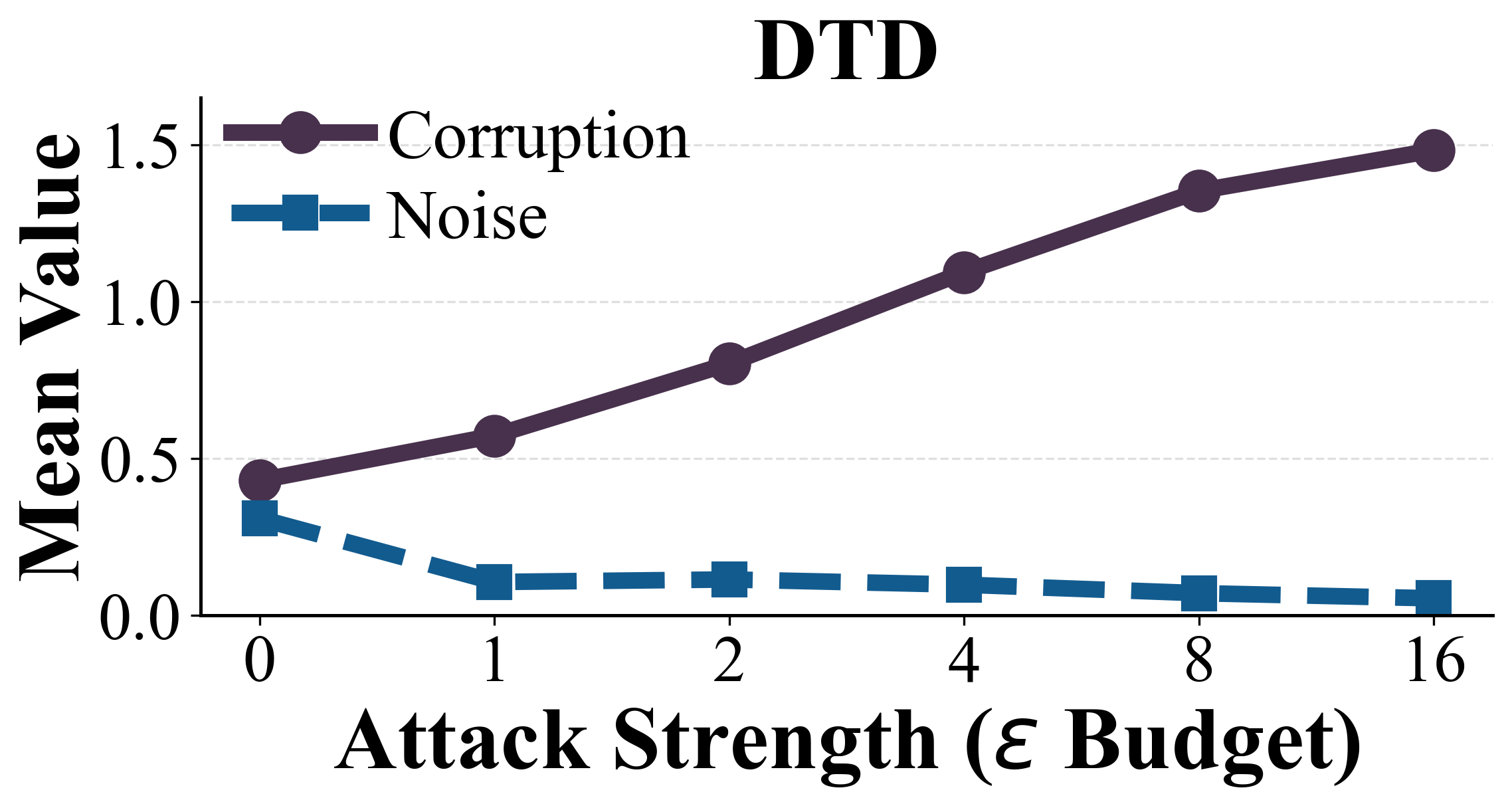}
    \vspace{-0.8em}
  \end{minipage}
  \hfill
  \begin{minipage}[t]{0.24\linewidth}
    \centering
    \includegraphics[width=\linewidth]{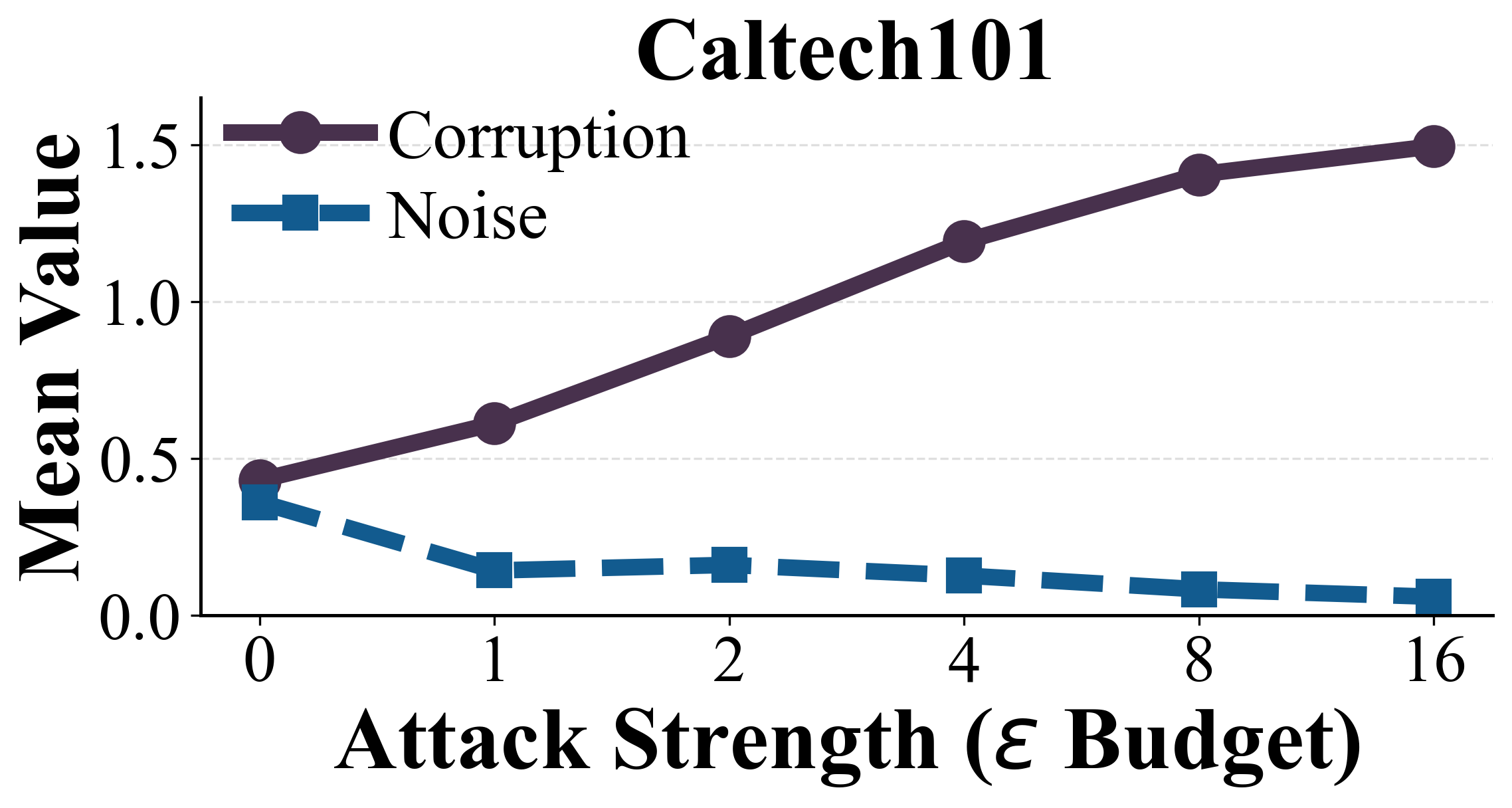}
    \vspace{-0.8em}
  \end{minipage}
  \hfill
  \begin{minipage}[t]{0.24\linewidth}
    \centering
    \includegraphics[width=\linewidth]{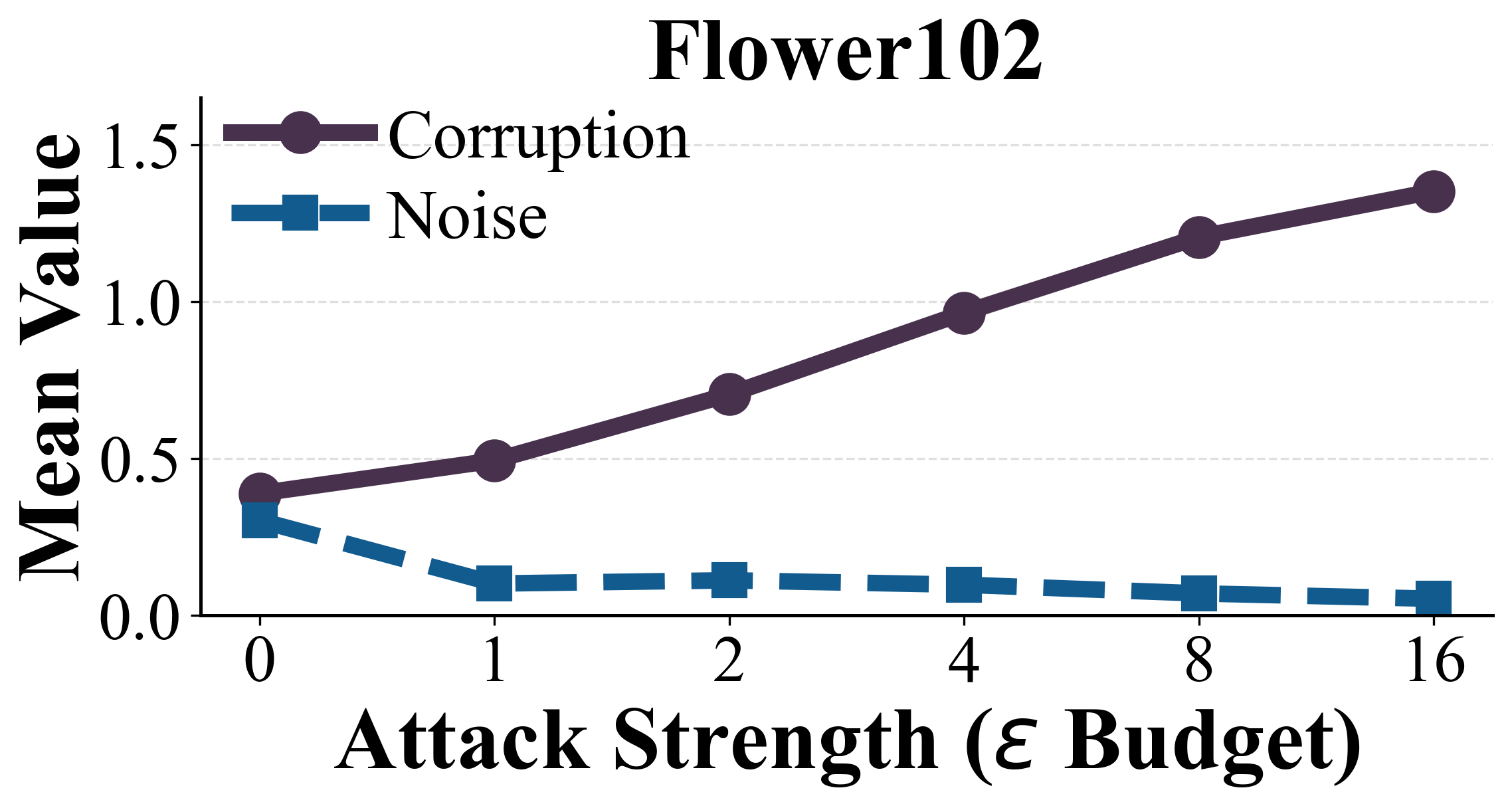}
    \vspace{-0.8em}
  \end{minipage}
  \hfill
  \begin{minipage}[t]{0.24\linewidth}
    \centering
    \includegraphics[width=\linewidth]{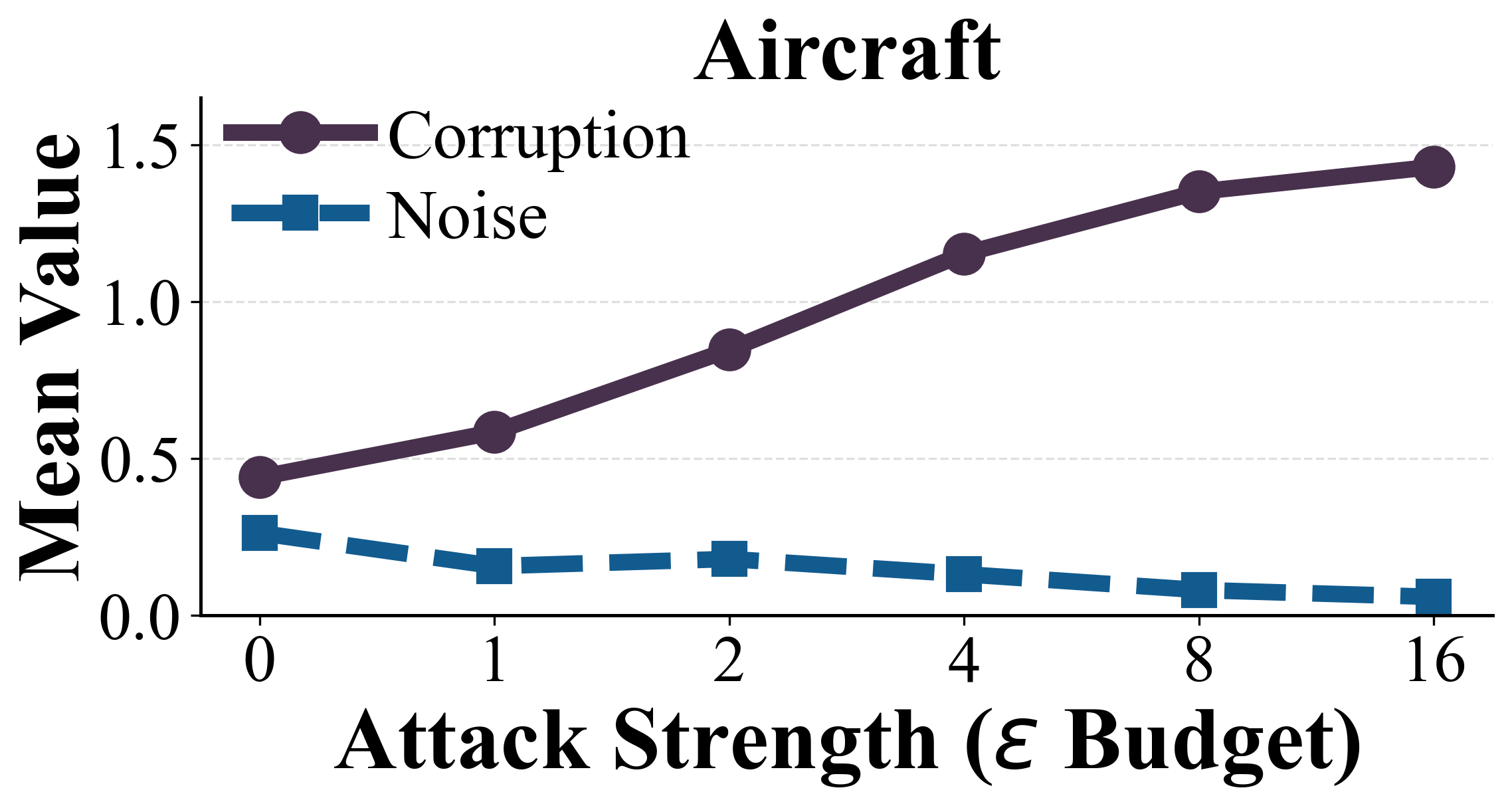}
    \vspace{-0.8em}
  \end{minipage}
  \\[0.7em]

  \begin{minipage}[t]{0.24\linewidth}
    \centering
    \includegraphics[width=\linewidth]{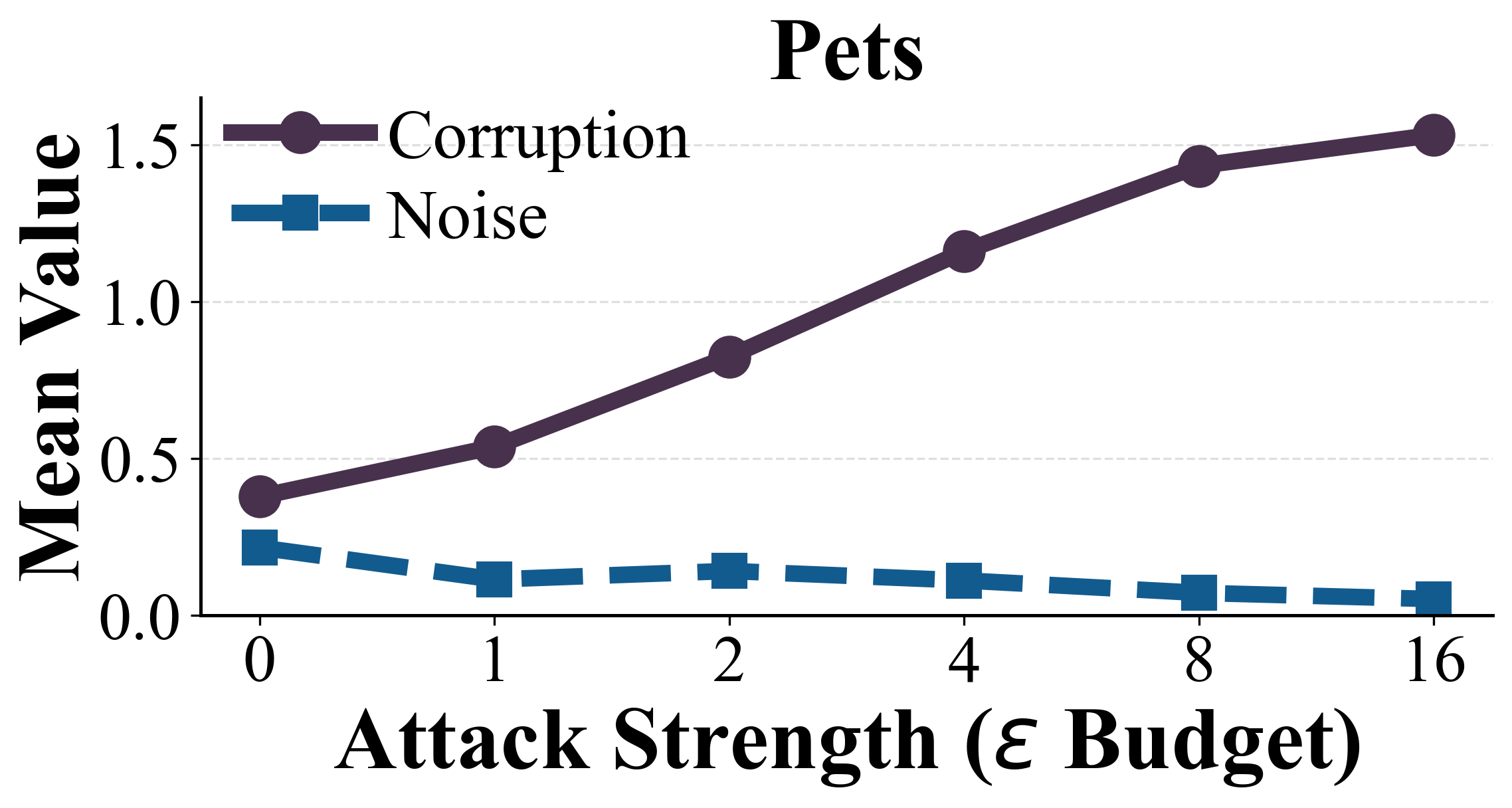}
    \vspace{-0.8em}
  \end{minipage}
  \hfill
  \begin{minipage}[t]{0.24\linewidth}
    \centering
    \includegraphics[width=\linewidth]{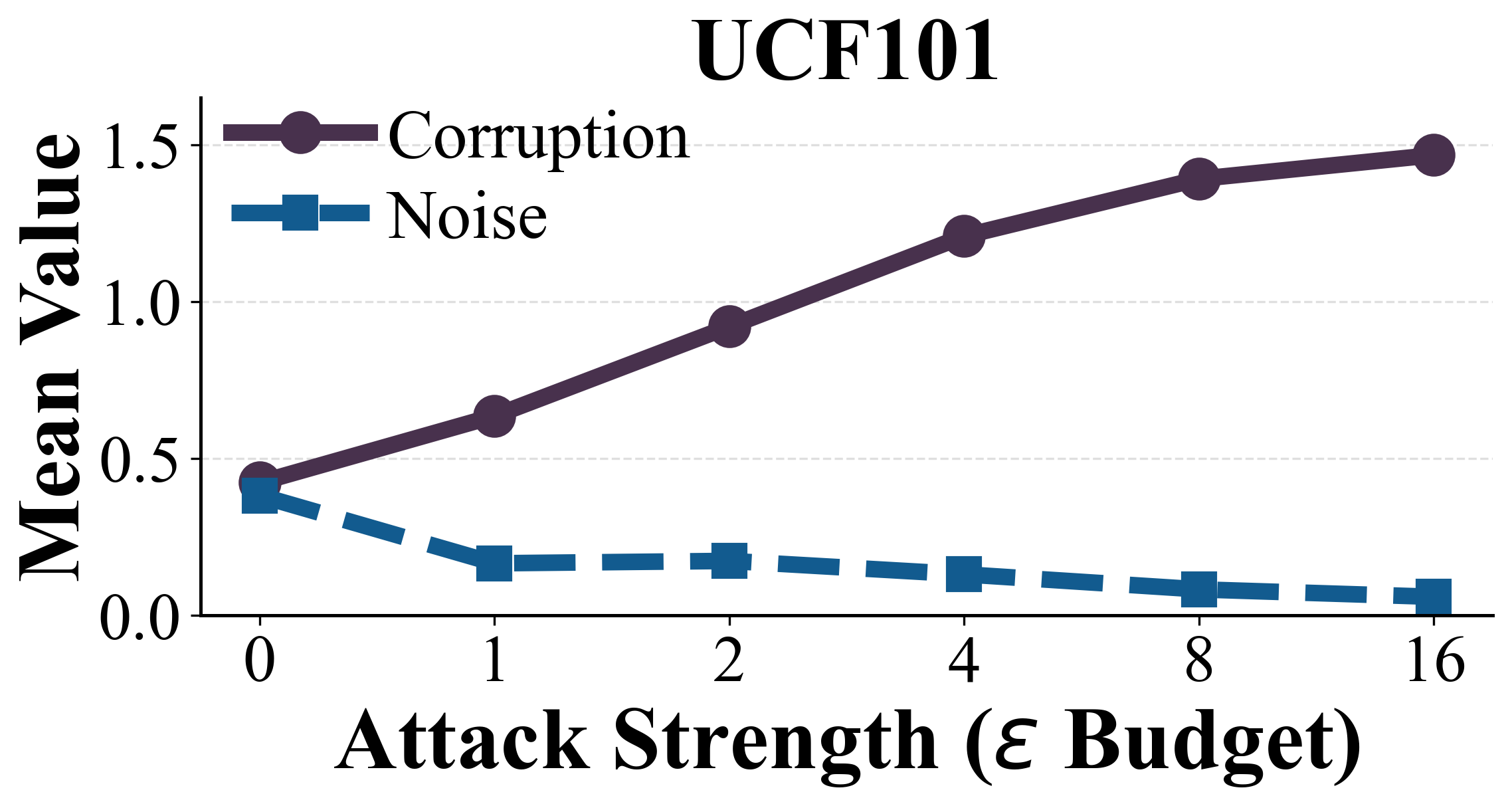}
    \vspace{-0.8em}
  \end{minipage}
  \hfill
  \begin{minipage}[t]{0.24\linewidth}
    \centering
    \includegraphics[width=\linewidth]{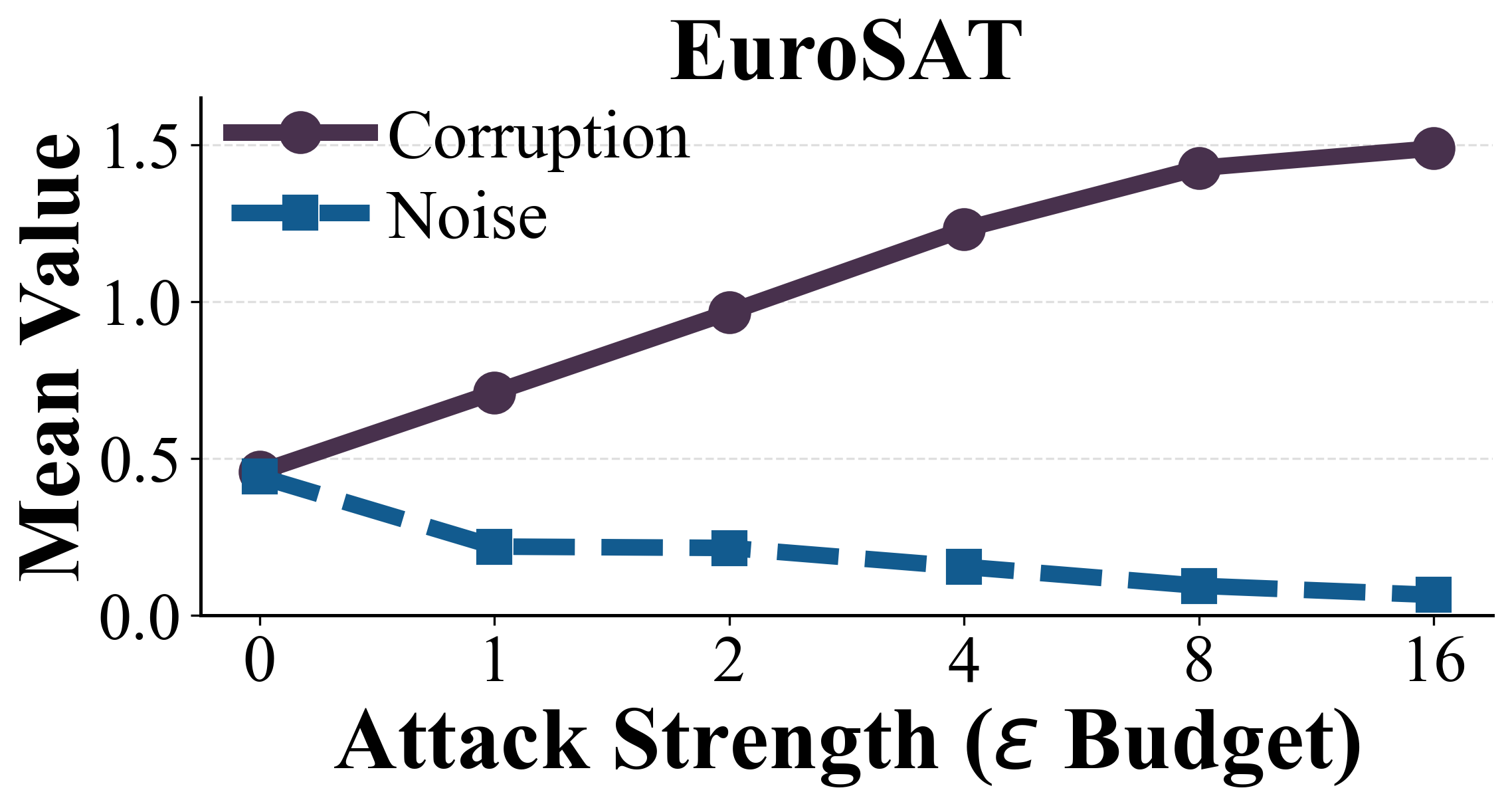}
    \vspace{-0.8em}
  \end{minipage}
  \hfill
  \begin{minipage}[t]{0.24\linewidth}
    \centering
    \includegraphics[width=\linewidth]{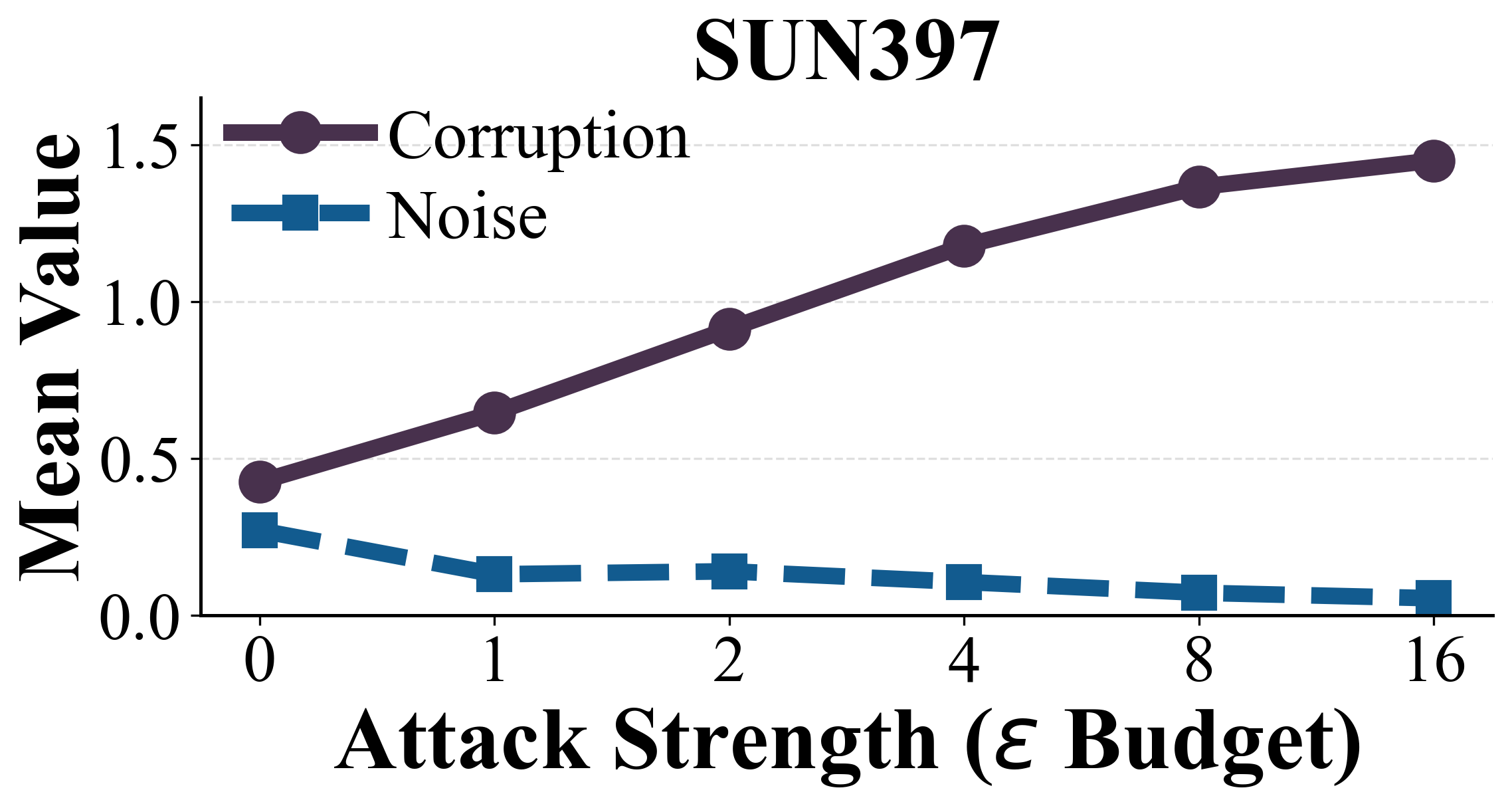}
    \vspace{-0.8em}
  \end{minipage}
  \\[0.7em]

  \begin{minipage}[t]{0.24\linewidth}
    \centering
    \includegraphics[width=\linewidth]{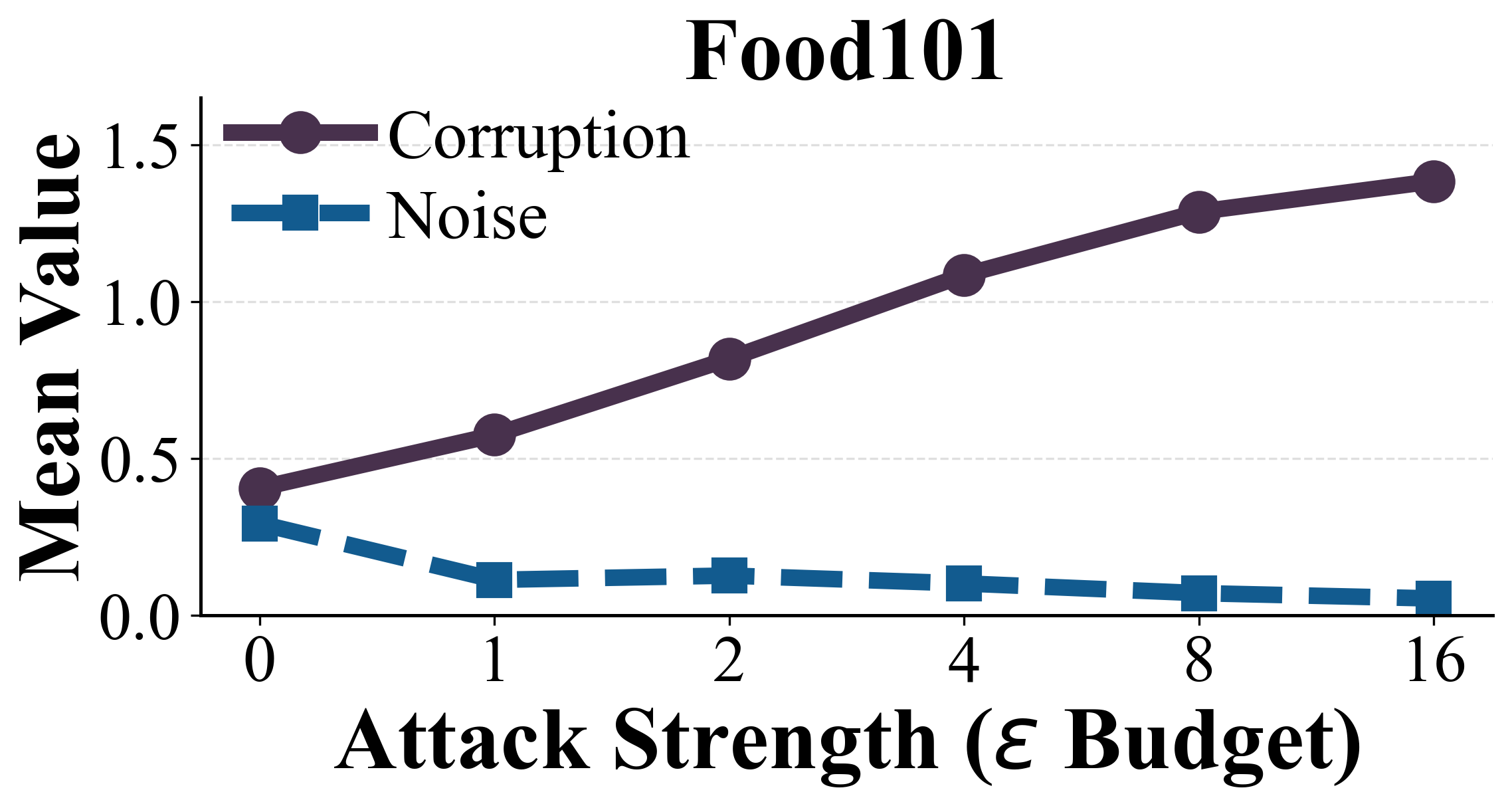}
    \vspace{-0.8em}
  \end{minipage}
  \hfill
  \begin{minipage}[t]{0.24\linewidth}
    \centering
    \includegraphics[width=\linewidth]{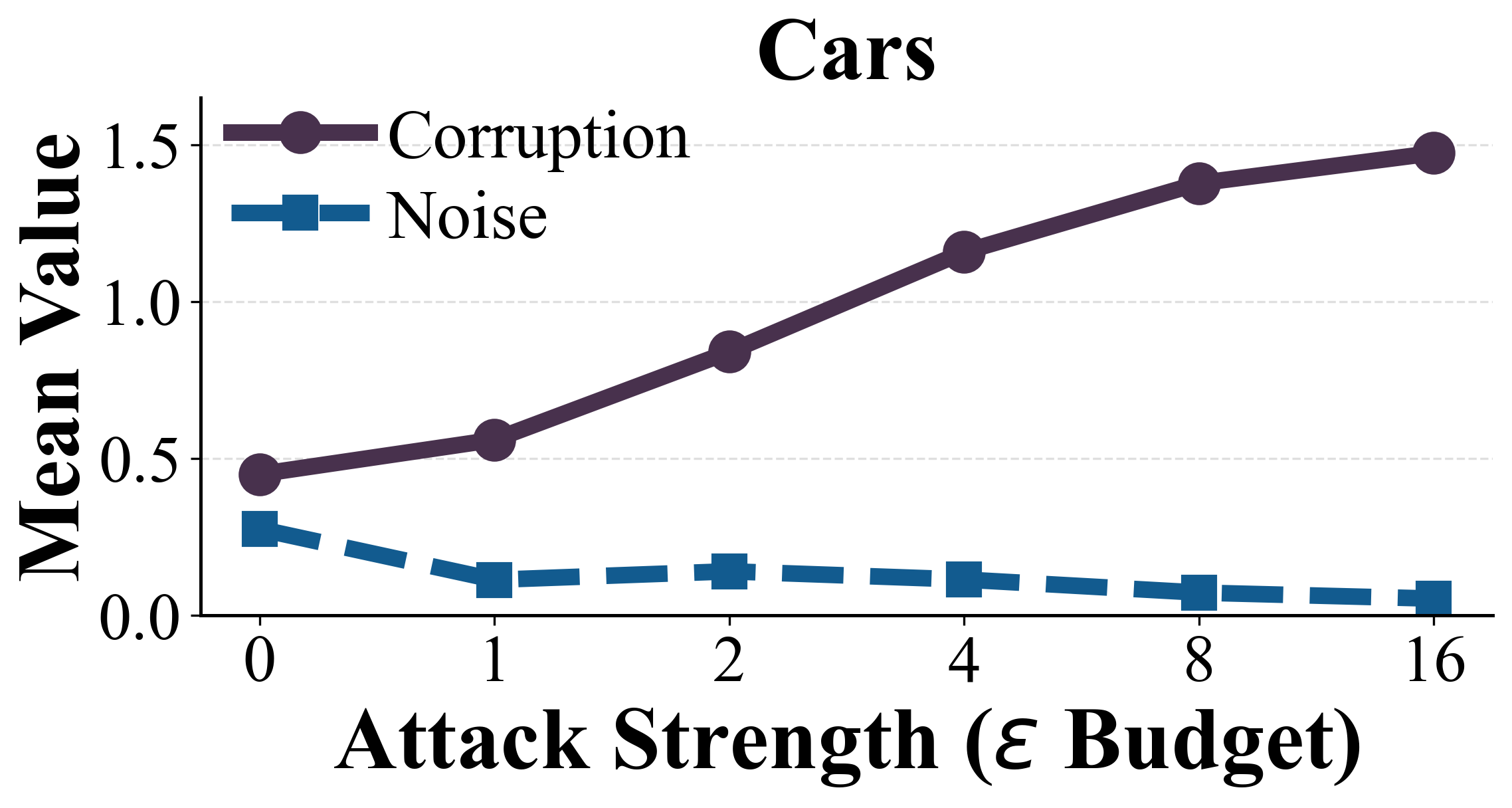}
    \vspace{-0.8em}
  \end{minipage}
  \hfill
  \begin{minipage}[t]{0.24\linewidth}
    \centering
    \includegraphics[width=\linewidth]{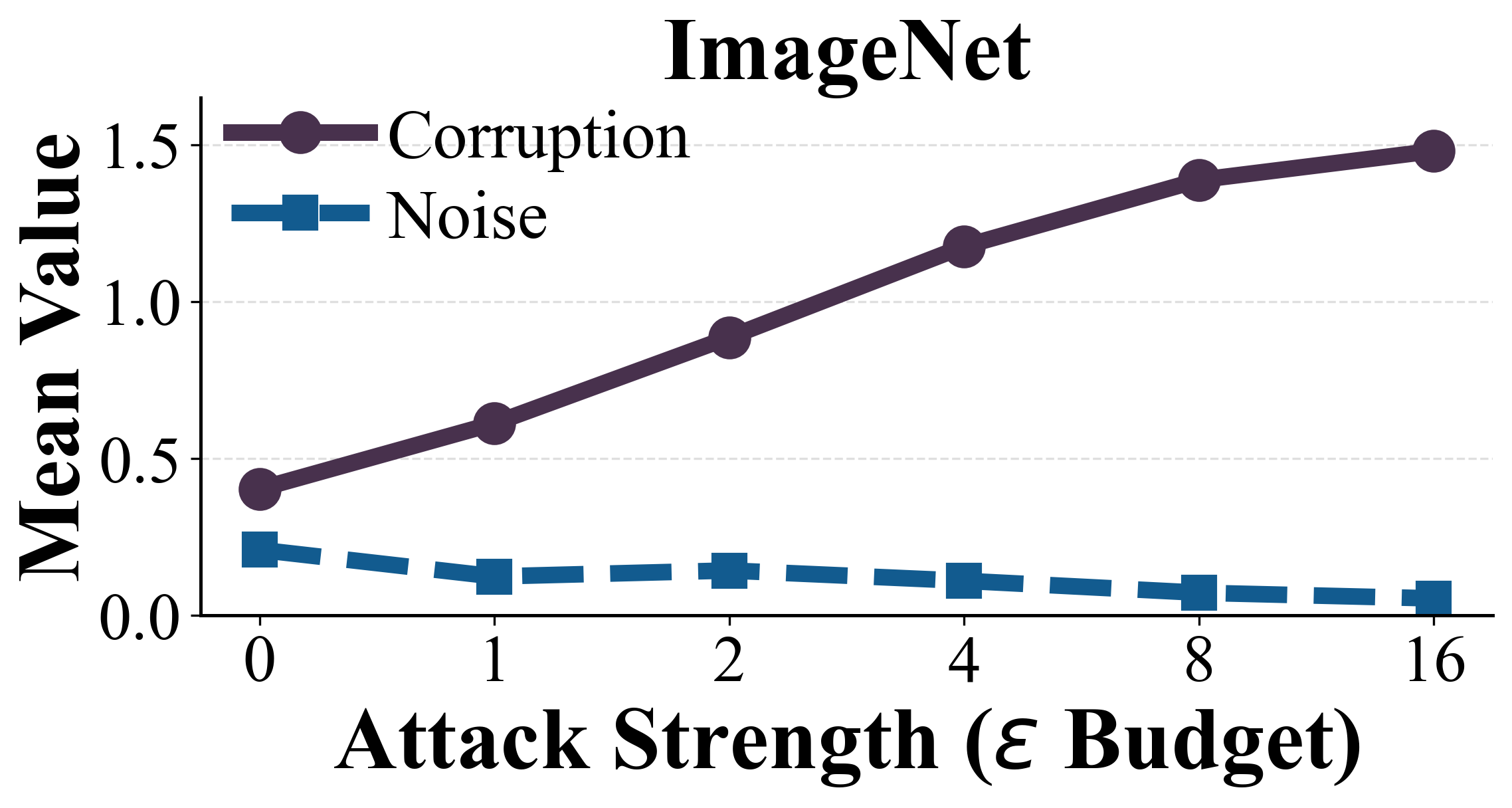}
    \vspace{-0.8em}
  \end{minipage}
  \hfill
  \begin{minipage}[t]{0.24\linewidth}
    \centering
    \includegraphics[width=\linewidth]{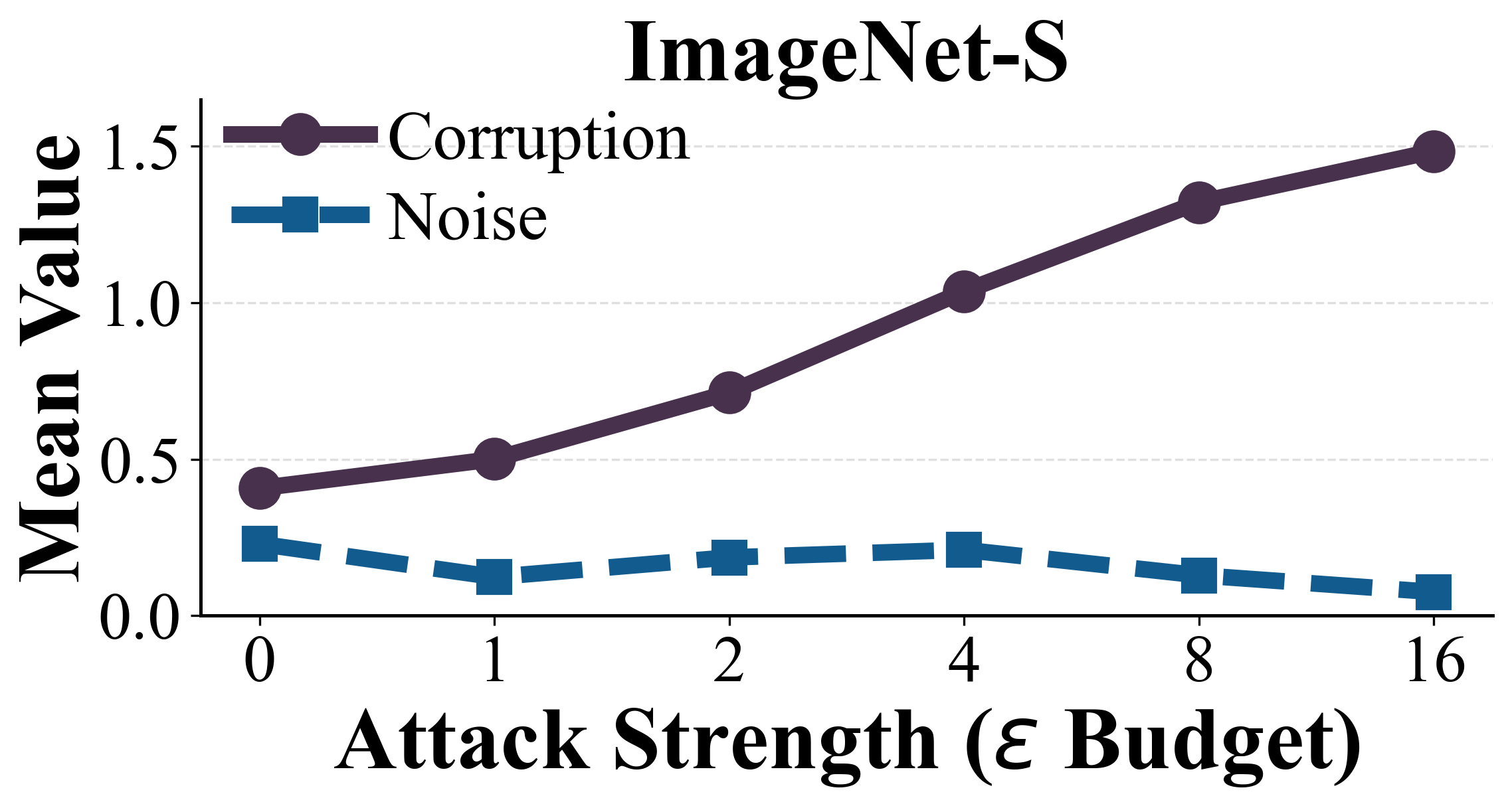}
    \vspace{-0.8em}
  \end{minipage}
  \\[0.7em]

  \begin{minipage}[t]{0.24\linewidth}
    \centering
    \includegraphics[width=\linewidth]{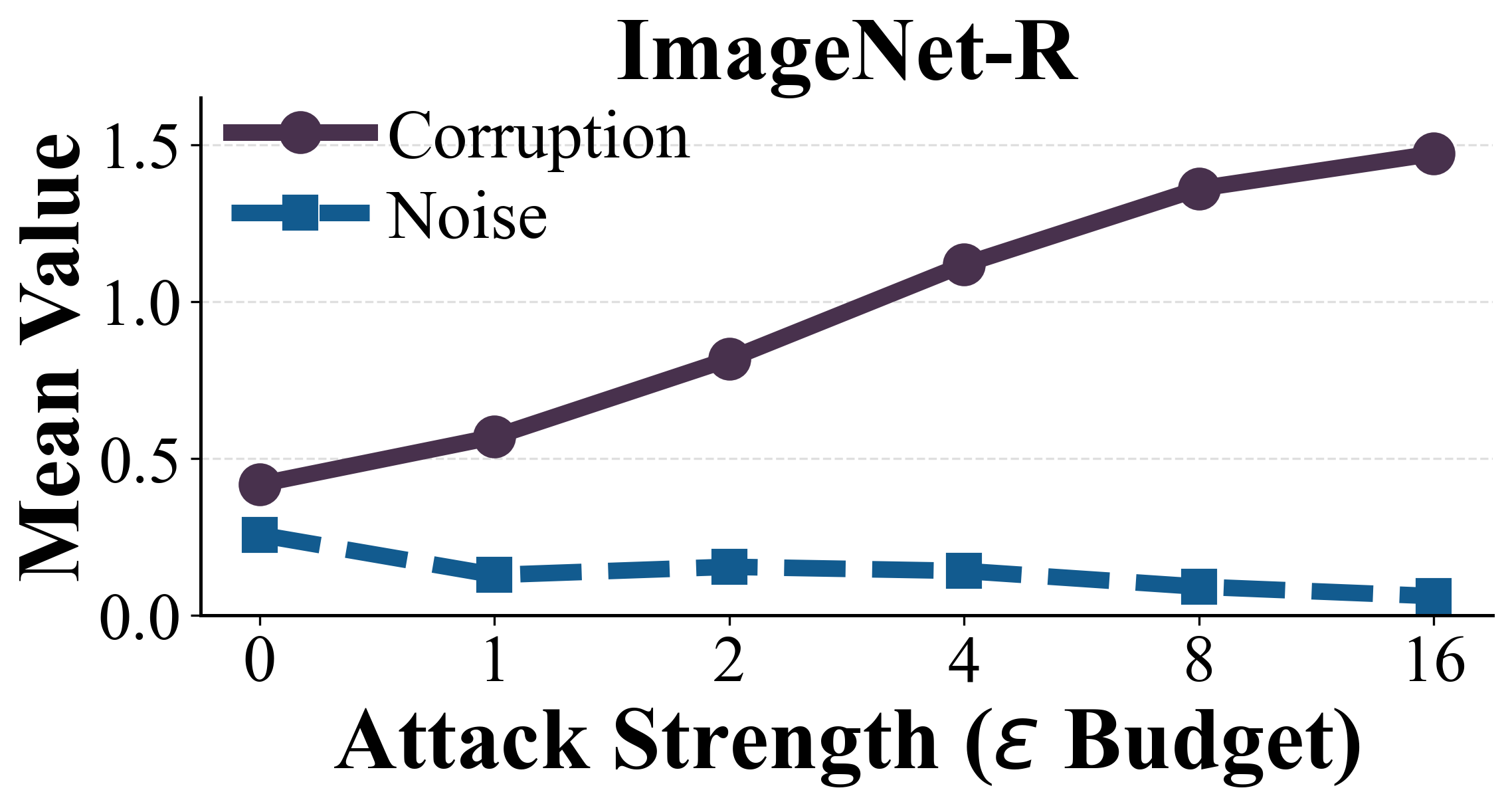}
    \vspace{-0.8em}
  \end{minipage}
  \hfill
  \begin{minipage}[t]{0.24\linewidth}
    \centering
    \includegraphics[width=\linewidth]{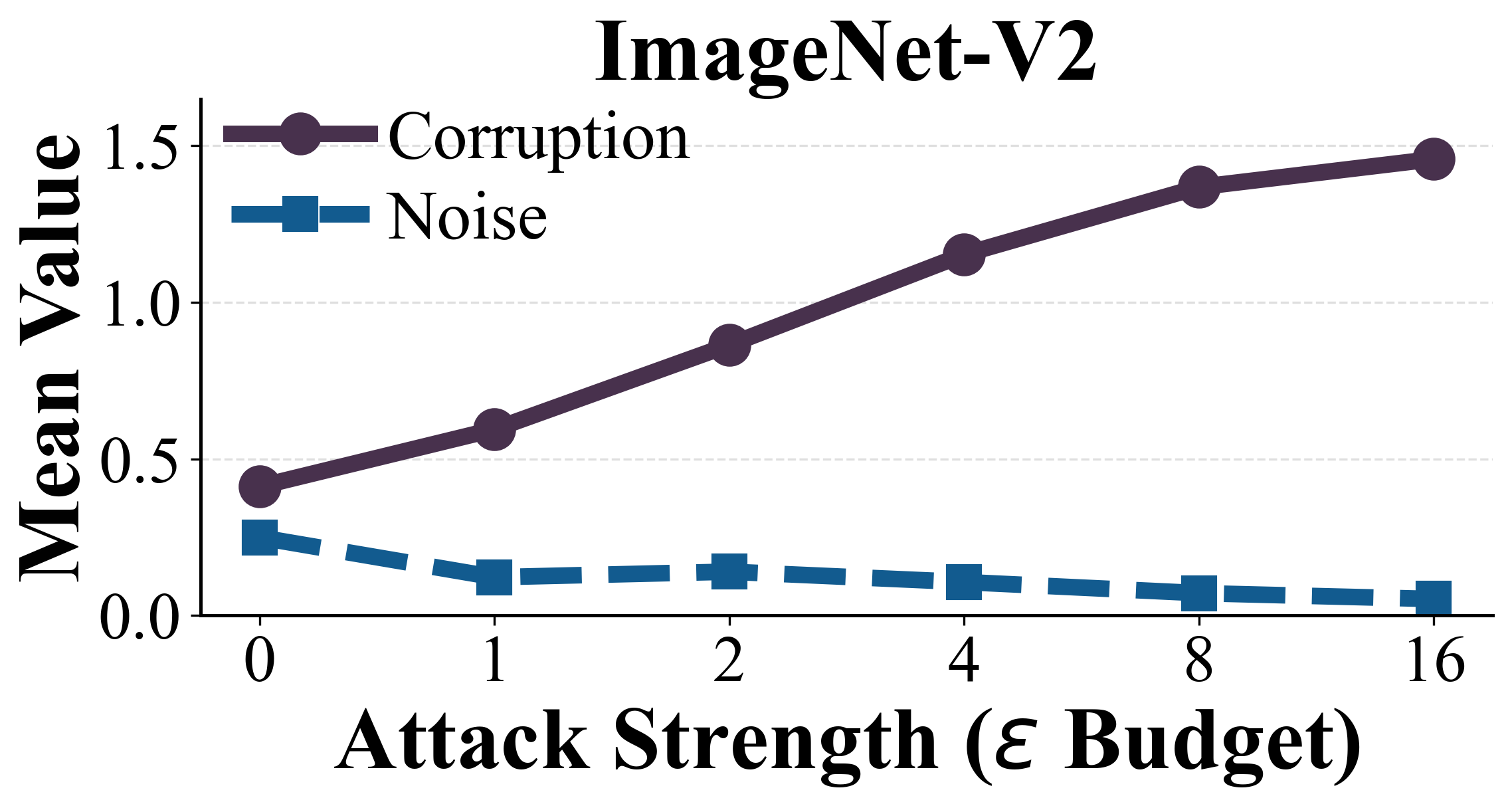}
    \vspace{-0.8em}
  \end{minipage}
  \hfill
  \begin{minipage}[t]{0.24\linewidth}
    \centering
    \includegraphics[width=\linewidth]{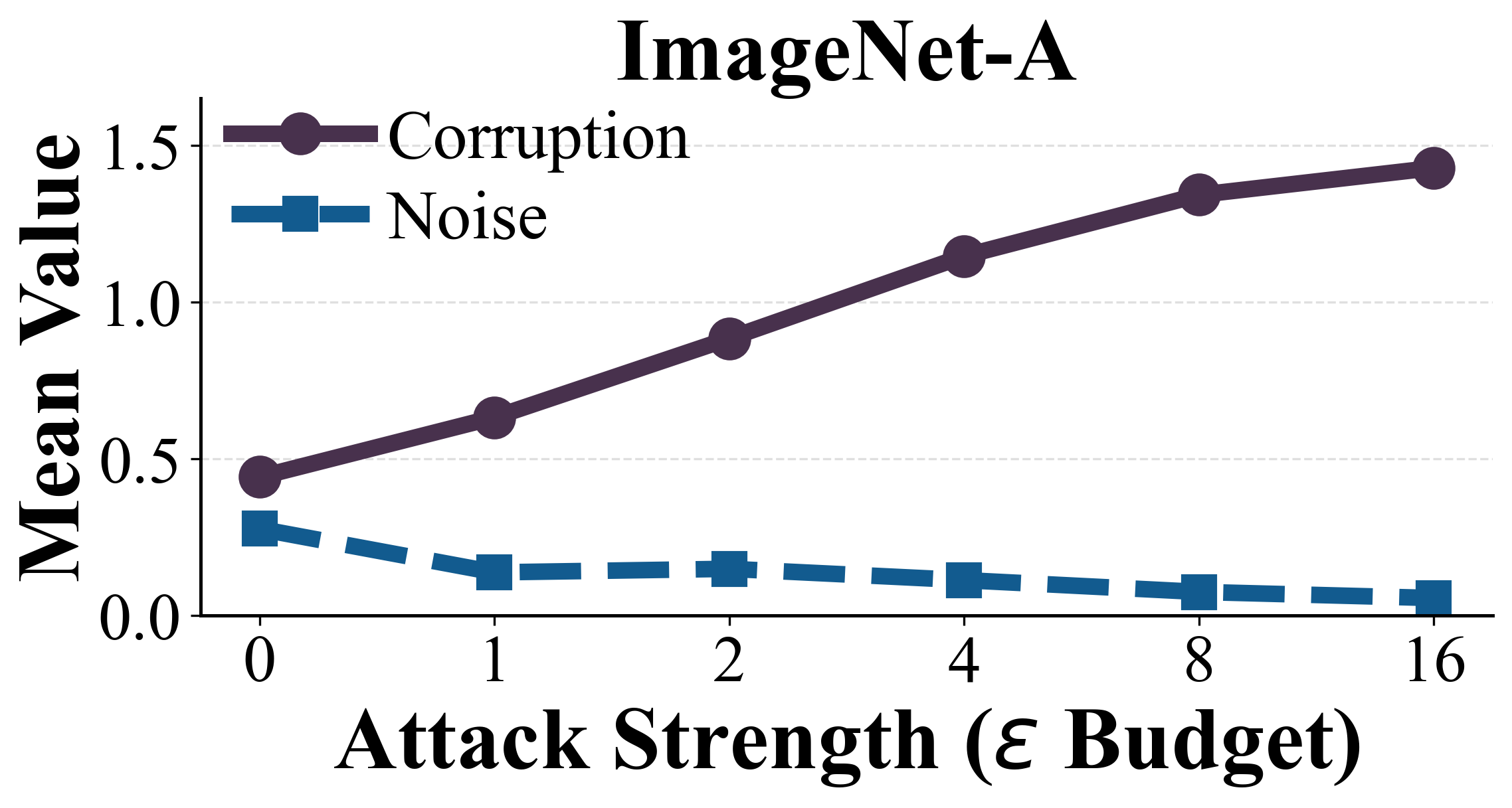}
    \vspace{-0.8em}
  \end{minipage}
  \hfill
  \begin{minipage}[t]{0.24\linewidth}
    \centering
    \includegraphics[width=\linewidth]{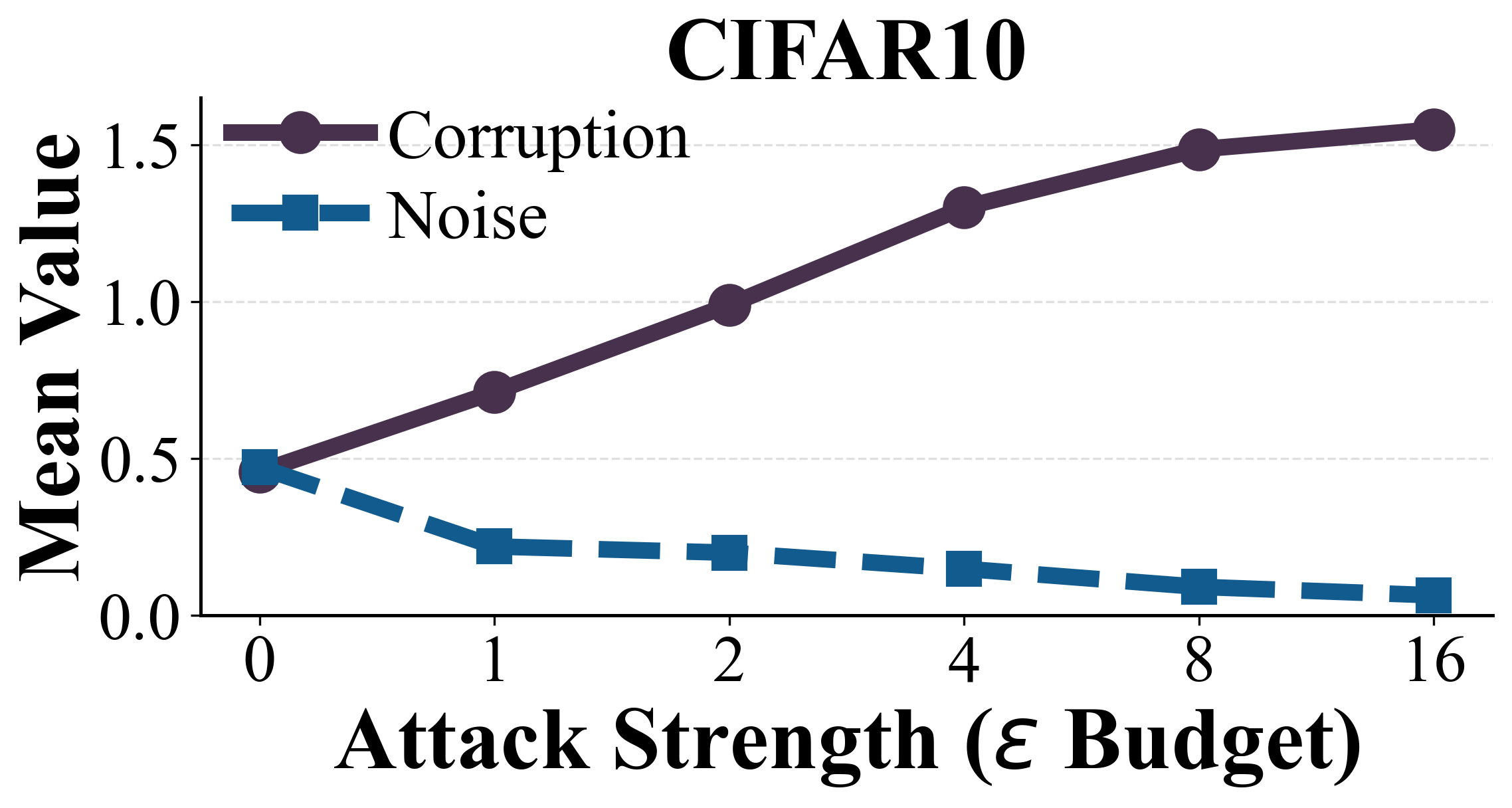}
    \vspace{-0.8em}
  \end{minipage}
  \\[0.7em]

  \begin{minipage}[t]{0.24\linewidth}
    \centering
    \includegraphics[width=\linewidth]{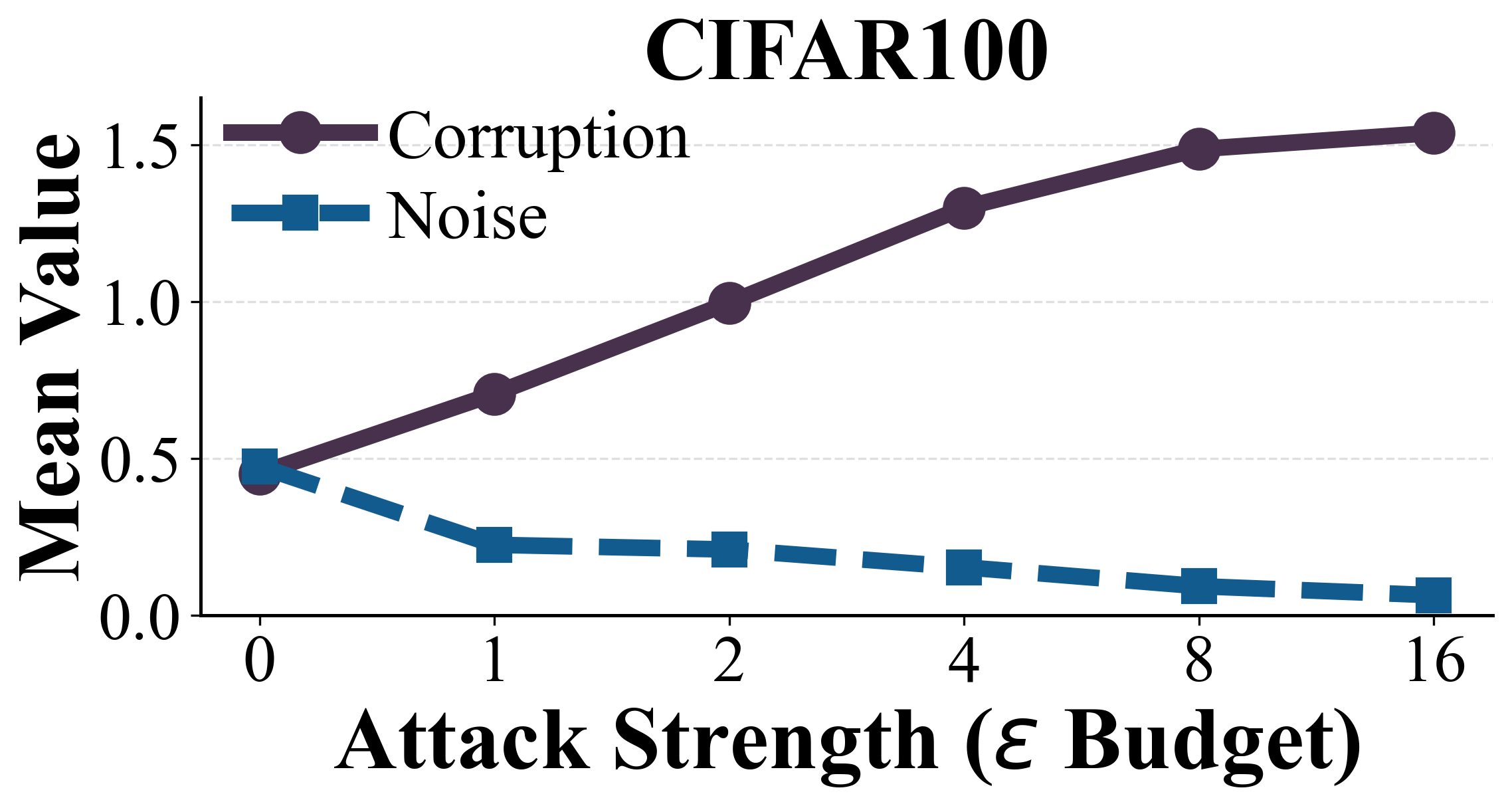}
    \vspace{-0.8em}
  \end{minipage}
  \hfill
  \begin{minipage}[t]{0.24\linewidth}
    \centering
    \includegraphics[width=\linewidth]{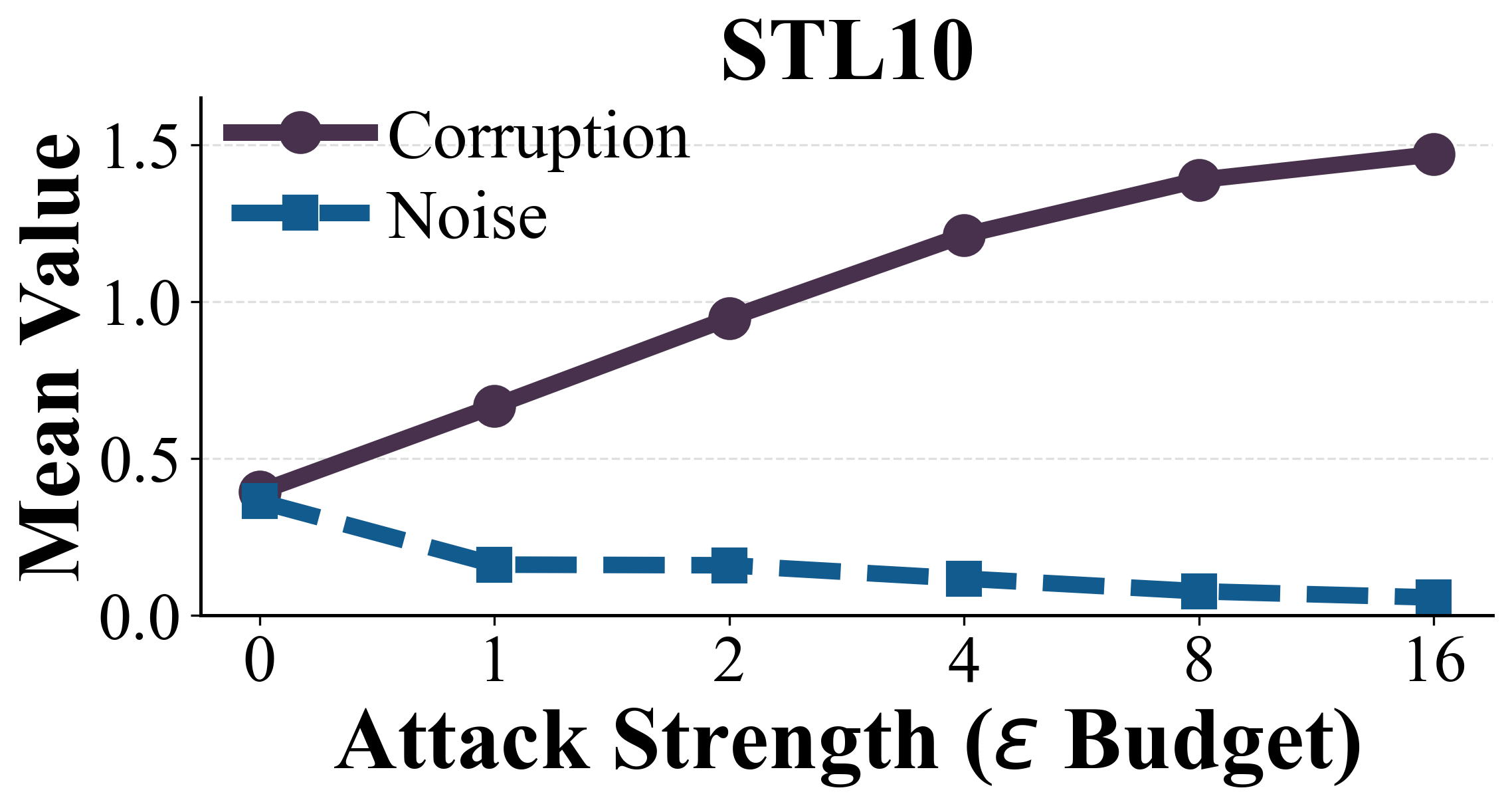}
    \vspace{-0.8em}
  \end{minipage}
  \hfill
  \begin{minipage}[t]{0.24\linewidth}
    \centering
    \includegraphics[width=\linewidth]{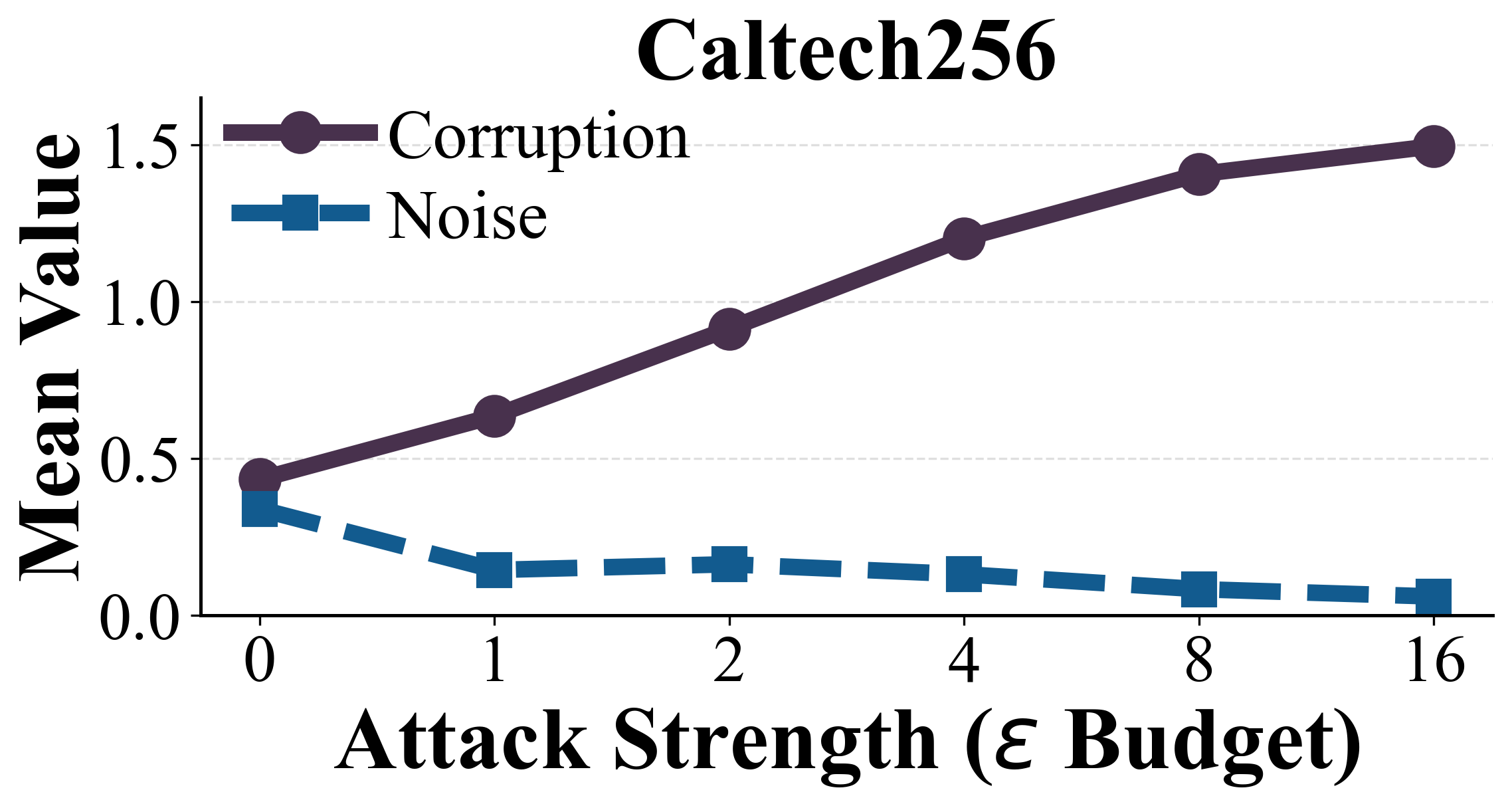}
    \vspace{-0.8em}
  \end{minipage}
  \hfill
  \begin{minipage}[t]{0.24\linewidth}
    \centering
    \includegraphics[width=\linewidth]{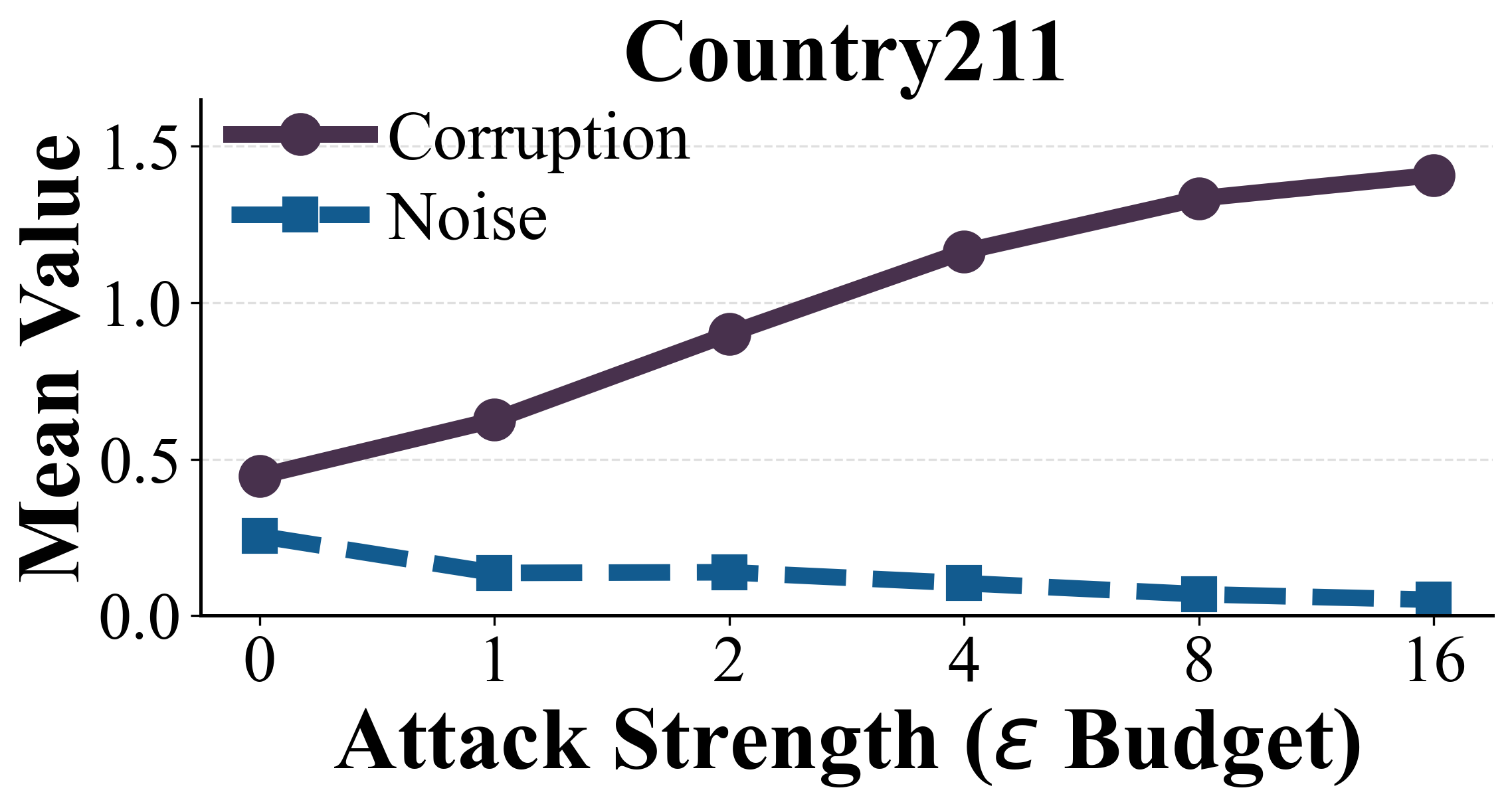}
    \vspace{-0.8em}
  \end{minipage}

  \vspace{-0.5em}
  \caption{
  Comparison of the mean corruption degree of our MAC and the mean noise-driven deviation of TTC~\cite{xing2025clip} under varying attack strengths $\epsilon^{(\mathrm{atk})}$ across 20 datasets.
  The corruption degree increases monotonically with stronger attacks and exhibits consistent trends across all datasets, whereas the noise-driven deviation shows non-monotonic behaviors.
  This demonstrates that our corruption degree provides a more reliable and stable indicator of corruption severity than the noise-driven deviation.}
  \label{fig:corruption_vs_noise_all}
  \vspace{-0.3em}
\end{figure*}

\paragraph{Attack settings.}
For all attack settings, perturbations are applied to image tensors normalized to the $[0,1]$ range, where the perturbation budget is expressed in the $[0,1]$ domain as $\epsilon^{(\mathrm{atk})}/255$.
We enable a random initialization and use one restart for the attack to ensure strong adversarial pressure.
The detailed configurations are as follows:

\begin{itemize}
\item \textbf{PGD attack.}
We use an $L_\infty$-bounded PGD attack with a perturbation budget of $\epsilon^{(\mathrm{atk})}=4$.
The step size is set to $\epsilon^{(\mathrm{atk})}/4$, following common practice.
We run $100$ iterations for CLIP ViT models and $10$ iterations for CLIP ResNet models.

\item \textbf{DI\textsuperscript{2}-FGSM attack.}  
We use an $L_\infty$-bounded DI\textsuperscript{2}-FGSM attack with the same $\epsilon^{(\mathrm{atk})}$, step size, and number of iterations as PGD.  
All diversity-transformation parameters follow standard settings, with a resize rate of $0.9$, diversity probability of $0.5$, and decay set to $0.0$.

\item \textbf{CW attack.}  
We adopt the $L_2$-bounded Carlini-Wagner attack with $c{=}3.0$, $\kappa{=}0$, $500$ optimization steps, and a learning rate of $0.01$.

\item \textbf{AutoAttack.}  
We use the $L_\infty$-bounded standard AutoAttack configuration with the same $\epsilon^{(\mathrm{atk})}$ as PGD, keeping all internal thresholds and settings at default values of their library.

\item \textbf{MAC-adaptive attack.}
To evaluate robustness under fully adaptive white-box threats, we design a pipeline-aware PGD attack that explicitly incorporates the MAC defense into the attack objective.
At each iteration, the attacker forms a perturbed image, generates $N$ views using the augmentation distribution $\mathcal{T}$, and then applies the multi-view guided counterattack with corruption-aware soft weighting to these views.
The attack maximizes the cross-entropy loss between the aggregated logits over the counterattacked views and the ground-truth label, thereby increasing the model's prediction error and effectively implementing an expectation-over-transformations (EOT) objective~\cite{athalye2018synthesizing}.
To backpropagate through MAC, including its non-differentiable operations, we employ a BPDA-style straight-through estimator~\cite{athalye2018obfuscated,bengio2013estimating}: the forward pass runs the full MAC pipeline, while the backward pass treats it as the identity transformation.
The perturbation $\delta$ is updated using an $L_\infty$-bounded PGD procedure with the same step size, budget, and number of iterations as in the standard PGD setting, resulting in a strictly defense-aware MAC-adaptive attack.

\end{itemize}

\begin{table}[t]
\caption{
Comparison of clean accuracy (Acc.) and adversarial robustness (Rob.) of TTC and our MAC across different architectures, such as CLIP RN50, ViT-B/32, ViT-B/16, and ViT-L/14, and counterattack perturbation budgets $\epsilon^{(\mathrm{ca})} \in \{4, 8, 16\}$. 
Results are averaged over ten fine-grained recognition datasets.
}
\centering
\small
\setlength{\tabcolsep}{7pt}
\resizebox{\linewidth}{!}{
\begin{tabular}{l|cc|c|cc|cc}
\toprule
 CLIP
& \multicolumn{2}{c|}{CLIP~\cite{radford2021learning}} 
& \multirow{2}{*}{$\epsilon^{(\mathrm{ca})}$} 
& \multicolumn{2}{c|}{TTC~\cite{xing2025clip}} 
& \multicolumn{2}{c}{MAC (Ours)} \\
Architecture & Acc. & \cellcolor{robgreen}Rob. &  & Acc. & \cellcolor{robgreen}Rob. & Acc. & \cellcolor{robgreen}Rob. \\
\midrule
\multirow{3}{*}{RN50}
 &    &  \cellcolor{robgreen}   & 4  & 52.0 & \cellcolor{robgreen}0.8 & \textbf{53.6} & \cellcolor{robgreen}\textbf{22.0} \\
 & 54.5  &   \cellcolor{robgreen}0.1  & 8  & 50.1 & \cellcolor{robgreen}2.9 & \textbf{53.4} & \cellcolor{robgreen}\textbf{34.8} \\
 &     &   \cellcolor{robgreen}  & 16 & 47.6 & \cellcolor{robgreen}23.9 & \textbf{53.3} & \cellcolor{robgreen}\textbf{38.7}\\
\midrule
\multirow{3}{*}{ViT-B/32}
 &     &   \cellcolor{robgreen}  & 4  & 56.6 & \cellcolor{robgreen}6.8 & \textbf{58.7} & \cellcolor{robgreen}\textbf{33.7} \\
 & 58.9   &   \cellcolor{robgreen}0.0  & 8  & 56.2 & \cellcolor{robgreen}3.9 & \textbf{58.7} & \cellcolor{robgreen}\textbf{45.2} \\
 &     &   \cellcolor{robgreen}  & 16 & 55.8 & \cellcolor{robgreen}1.3 & \textbf{58.6} & \cellcolor{robgreen}\textbf{46.2} \\
\midrule
\multirow{3}{*}{ViT-B/16}
 &     &   \cellcolor{robgreen}  & 4  & 60.5 & \cellcolor{robgreen}4.8 & \textbf{62.3} & \cellcolor{robgreen}\textbf{48.7} \\
 & 62.7  &   \cellcolor{robgreen}0.0  & 8  & 59.8 & \cellcolor{robgreen}0.2 & \textbf{62.3} & \cellcolor{robgreen}\textbf{58.6} \\
 &     &  \cellcolor{robgreen}  & 16 & 58.7 & \cellcolor{robgreen}11.1 & \textbf{62.3} & \cellcolor{robgreen}\textbf{58.8} \\
 \midrule
\multirow{3}{*}{ViT-L/14}
 &     &   \cellcolor{robgreen}  & 4  & 68.6 & \cellcolor{robgreen}4.6 & \textbf{69.5} & \cellcolor{robgreen}\textbf{57.3} \\
 & 69.9  &   \cellcolor{robgreen}0.0  & 8  & 68.5 & \cellcolor{robgreen}0.8 & \textbf{69.4} & \cellcolor{robgreen}\textbf{64.3} \\
 &     &  \cellcolor{robgreen}  & 16 & 67.2 & \cellcolor{robgreen}25.8 & \textbf{69.3} & \cellcolor{robgreen}\textbf{64.3} \\
\bottomrule
\end{tabular}}
\label{tab:arch_ttceps_comparison}
\end{table}

\section{Experimental Results}
\label{suppl-sec:exp}

\paragraph{Comparisons with adversarial fine-tuning methods.}
We compare our method with adversarial fine-tuning approaches and test-time defenses under strong PGD attack settings reported in~\cite{xing2025clip}.
All performance values for the compared methods are taken from the supplementary material of~\cite{xing2025clip}, where further implementation details can also be found.
As shown in~\cref{tab:adv_finetuning_comparisons}, adversarial fine-tuning approaches such as CLIP-FT~\cite{xing2025clip}, TeCoA~\cite{mao2022understanding}, PMG-AFT~\cite{wang2024pre}, and FARE~\cite{schlarmann2024robust} offer moderate robustness gains but incur substantial degradation in clean accuracy and still yield limited adversarial performance.
Test-time defense methods, including RN~\cite{xing2025clip}, TTE~\cite{perez2021enhancing}, Anti-adv~\cite{alfarra2022combating}, and HD~\cite{wu2021attacking},  preserve high clean accuracy but provide only modest robustness improvements.

In contrast, MAC achieves the highest robustness across all datasets in the benchmark, reaching \textbf{36.83\%} average adversarial accuracy while maintaining competitive clean accuracy (\textbf{66.55\%}).
These results demonstrate that MAC achieves a substantially better clean-robustness trade-off than both adversarial fine-tuning and prior test-time defense methods, without requiring any model fine-tuning.

\paragraph{Comparisons with TTC across additional datasets.}
As shown in~\cref{tab:adv_finetuning_comparisons}, MAC also surpasses TTC on the five datasets, improving the average adversarial accuracy from \textbf{23.93\%} to \textbf{36.83\%}.
When these results are considered together with the broader evaluations in \cref{tab:vit_finegrained_sota_comparisons} and \cref{tab:vit_imagenet_sota_comparisons}, MAC is consistently superior to TTC across all \emph{20 datasets} evaluated.
Taken collectively, these findings establish MAC as a strictly stronger counterattack mechanism than TTC across fine-grained, large-scale, and OOD benchmarks.

\paragraph{Stability across architectures and perturbation budgets.}
We analyze the behavior of TTC and MAC across different architectures and counterattack perturbation budgets $\epsilon^{(\mathrm{ca})}$, as shown in \cref{tab:arch_ttceps_comparison}. 
While TTC exhibits inconsistent behavior across perturbation levels, where its clean accuracy and adversarial robustness fluctuate significantly within the same architecture as $\epsilon^{(\mathrm{ca})}$ varies, MAC demonstrates both stable clean performance and steadily increasing robustness as the $\epsilon^{(\mathrm{ca})}$ increases. 
This suggests that TTC's noise-driven hard gating mechanism struggles to reliably distinguish between clean and adversarial inputs, leading to unstable accuracy and robustness. 
In contrast, MAC leverages the estimated corruption degree as a reliable indicator of corruption severity and incorporates it into a corruption-aware soft weighting scheme, which adaptively handles various input conditions, including clean, weakly perturbed, and strongly perturbed images.
As a result, MAC achieves consistently higher and more stable performance than TTC across various architectures and counterattack strengths.
We further investigate the underlying cause of this stability gap in the following paragraph by directly comparing MAC's corruption degree with TTC's noise-driven deviation.

\paragraph{Comparing the corruption degree of MAC and the noise-driven deviation of TTC.}
For each image, we use the \emph{corruption degree} as an indicator of corruption severity to adaptively scale the counterattack intensity.
Images with a degree above the threshold receive proportionally stronger counterattacks, while those below the threshold are only lightly updated.
In contrast, TTC~\cite{xing2025clip} employs a \emph{noise-driven deviation} as a binary trigger: only images with deviation below a fixed threshold are counterattacked.

To examine how faithfully each metric reflects actual corruption, we compare the behaviors of corruption degree and noise-driven deviation across 20 datasets under varying attack strengths, as shown in \cref{fig:corruption_vs_noise_all}.
The corruption degree of MAC increases \emph{monotonically} with attack strength and exhibits consistently similar curve shapes across all datasets, demonstrating that it serves as a stable and reliable corruption-severity indicator.
In contrast, the noise-driven deviation of TTC shows strongly \emph{non-monotonic} trends.
As the attack strength increases, the deviation does not consistently decrease; instead, it may rise, fall, or plateau depending on the dataset.
This instability is especially observed in distribution-shifted benchmarks such as ImageNet-S, where the deviation increases from $\epsilon^{(\mathrm{atk})}{=}1$ to $\epsilon^{(\mathrm{atk})}{=}4$, but then decreases again at $\epsilon^{(\mathrm{atk})}{=}8$.
Such irregular patterns indicate that noise-driven deviation fails to reliably measure corruption severity.

This discrepancy arises because noise-driven deviation of TTC is insensitive to structured, geometric, and semantic variations introduced by adversarial perturbations, whereas our augmentation-based corruption degree captures these structural changes by leveraging affine transformations, blur, and color jitter.
This fundamental difference directly explains the stability gap observed in \cref{tab:arch_ttceps_comparison}: MAC's corruption degree provides a reliable signal that supports stable behavior across architectures and counterattack perturbation budgets, whereas TTC's deviation leads to unstable clean accuracy and robustness.
Consequently, incorporating corruption degree into MAC yields substantial robustness improvements (\textbf{+37.7\%} on average) compared to incorporating noise-driven deviation into MAC, as validated by the ablation results in \cref{fig:ablation_masking}.

\begin{figure}[t!]
  \centering
  \includegraphics[width=0.9\linewidth]{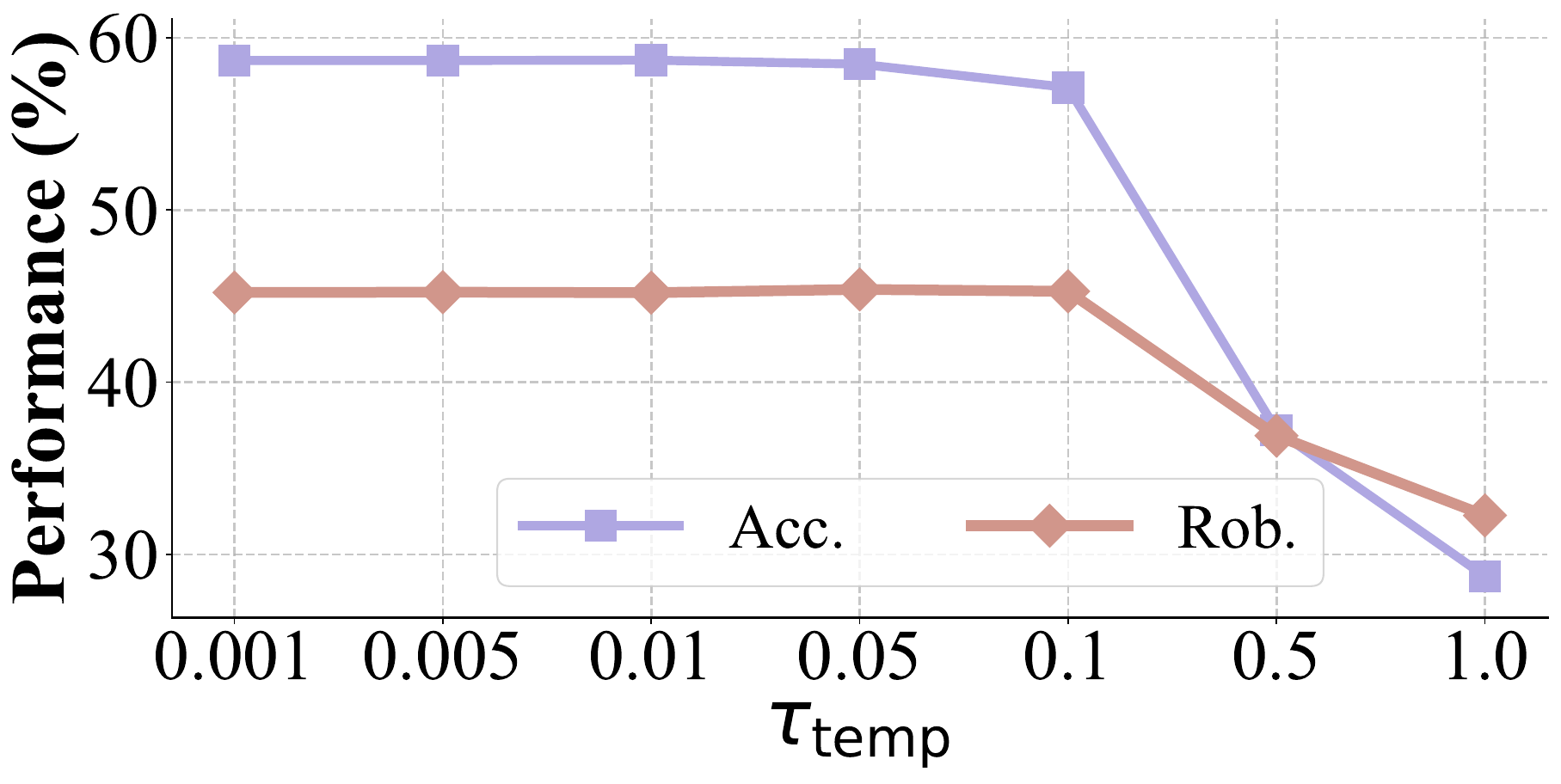}
  \caption{Analysis of the temperature parameter $\tau_{\mathrm{temp}}$, averaged over ten fine-grained recognition datasets. Moderate temperature values yield stable and near-optimal performance, while large temperatures reduce adaptivity and degrade overall performance.}
\label{fig:tau-temp}
\end{figure}

\begin{table*}[t]
\caption{
Evaluation of various test-time adaptation methods on eight fine-grained recognition datasets using CLIP-ViT-L/14, reporting both clean accuracy (Acc.) and adversarial accuracy under PGD-100 attack with $\epsilon$ = 4.0 (Rob.). The highest score is highlighted in bold. $\Delta$ Rob. represents the robust accuracy gain of our method over the best existing tuning-free method.
}
\centering
\small
\resizebox{0.9\textwidth}{!}{
\setlength{\tabcolsep}{6.5pt}
\begin{tabular}{lll|cccccccc|c}
\toprule
Category & Method & Metric & Caltech101 & DTD & Flower102 & Pets & UCF101 & Aircraft & EuroSAT & Cars & Average \\
\midrule
\multirow{2}{*}{Baseline} 
  & \multirow{2}{*}{CLIP~\cite{radford2021learning}} 
  & Acc. & 95.2 & 52.4 & \textbf{76.2} & 93.1 & 73.7 & 30.0 & \textbf{55.1} & 76.8 & \textbf{69.1} \\
  &  & \cellcolor{robgreen}Rob. 
        & \cellcolor{robgreen}0.1 
        & \cellcolor{robgreen}0.0 
        & \cellcolor{robgreen}0.0 
        & \cellcolor{robgreen}0.0 
        & \cellcolor{robgreen}0.0 
        & \cellcolor{robgreen}0.0 
        & \cellcolor{robgreen}0.0 
        & \cellcolor{robgreen}0.0 
        & \cellcolor{robgreen}0.0 \\
\midrule
\multirow{2}{*}{Tuning-based}
  & \multirow{2}{*}{R\mbox{-}TPT~\cite{sheng2025r}} 
  & Acc. & 95.7 & \textbf{54.0} & \textbf{76.2} & \textbf{93.7} & 74.3 & 31.7 & 44.3 & 77.2 & 68.4 \\
  &  & \cellcolor{robgreen}Rob. 
        & \cellcolor{robgreen}88.2 
        & \cellcolor{robgreen}38.0 
        & \cellcolor{robgreen}55.6 
        & \cellcolor{robgreen}72.9 
        & \cellcolor{robgreen}55.6 
        & \cellcolor{robgreen}17.2 
        & \cellcolor{robgreen}20.4 
        & \cellcolor{robgreen}49.1 
        & \cellcolor{robgreen}49.6 \\
\midrule
\multirow{8}{*}{Tuning-free} 
  & \multirow{2}{*}{MTA~\cite{zanella2024test}} 
  & Acc. & \textbf{95.8} & 53.4 & 76.1 & \textbf{93.7} & \textbf{74.7} & \textbf{32.7} & 47.8 & \textbf{78.4} & \textbf{69.1} \\
  &  & \cellcolor{robgreen}Rob. 
        & \cellcolor{robgreen}83.1 
        & \cellcolor{robgreen}27.2 
        & \cellcolor{robgreen}44.2 
        & \cellcolor{robgreen}64.9 
        & \cellcolor{robgreen}47.5 
        & \cellcolor{robgreen}8.0 
        & \cellcolor{robgreen}7.5 
        & \cellcolor{robgreen}36.6 
        & \cellcolor{robgreen}39.9 \\
        \cmidrule{2-12}
  & \multirow{2}{*}{TTC~\cite{xing2025clip}} 
  & Acc. & 93.5 & 51.0 & 74.5 & 92.2 & 72.7 & 27.9 & 45.9 & 74.6 & 66.5 \\
  &  & \cellcolor{robgreen}Rob. 
        & \cellcolor{robgreen}9.9 
        & \cellcolor{robgreen}5.3 
        & \cellcolor{robgreen}6.9 
        & \cellcolor{robgreen}6.4 
        & \cellcolor{robgreen}2.5 
        & \cellcolor{robgreen}0.4 
        & \cellcolor{robgreen}0.1 
        & \cellcolor{robgreen}3.2 
        & \cellcolor{robgreen}4.3 \\ \cmidrule{2-12}
  & \multirow{2}{*}{MAC (Ours)} 
  & Acc. & 94.1 & 52.1 & 75.0 & 93.6 & 73.1 & 28.0 & 47.5 & 74.6 & 67.3 \\
  &  & \cellcolor{robgreen}Rob. 
        & \cellcolor{robgreen}\textbf{92.7} 
        & \cellcolor{robgreen}\textbf{45.3} 
        & \cellcolor{robgreen}\textbf{70.4} 
        & \cellcolor{robgreen}\textbf{87.0} 
        & \cellcolor{robgreen}\textbf{68.7} 
        & \cellcolor{robgreen}\textbf{31.3} 
        & \cellcolor{robgreen}\textbf{25.6} 
        & \cellcolor{robgreen}\textbf{69.8}     
        & \cellcolor{robgreen}\textbf{61.4} \\\cmidrule{2-12}
  & \multicolumn{2}{l|}{\cellcolor{robgreen}$\Delta$ Rob.} 
    & \cellcolor{robgreen}\textcolor{teal}{\textbf{+9.6}}
    & \cellcolor{robgreen}\textcolor{teal}{\textbf{+18.1}}
    & \cellcolor{robgreen}\textcolor{teal}{\textbf{+26.2}}
    & \cellcolor{robgreen}\textcolor{teal}{\textbf{+22.1}}
    & \cellcolor{robgreen}\textcolor{teal}{\textbf{+21.2}}
    & \cellcolor{robgreen}\textcolor{teal}{\textbf{+23.3}}
    & \cellcolor{robgreen}\textcolor{teal}{\textbf{+18.1}}
    & \cellcolor{robgreen}\textcolor{teal}{\textbf{+33.2}}
    & \cellcolor{robgreen}\textcolor{teal}{\textbf{+21.5}} \\
\bottomrule
\end{tabular}}
\label{tab:vitl14_finegrained_sota_comparisons}
\end{table*}

\begin{table}[t]
\caption{Effect of removing each augmentation for multi-view generation. Results are averaged over ten fine-grained recognition datasets.}
\centering
\small
\resizebox{0.8\linewidth}{!}{
\setlength{\tabcolsep}{11pt}
\begin{tabular}{l|cc}
\toprule
Augmentation Configuration & Acc. & Rob. \\
\midrule
MAC w/o Affine Transform  & 58.6 &  6.0 \\
MAC w/o Color Jitter      & 58.6 & 44.9 \\
MAC w/o Gaussian Noise    & 58.4 & 44.7 \\
MAC w/o Gaussian Blur     & \textbf{58.8} & 42.3 \\
\midrule
MAC     & 58.7 & \textbf{45.2} \\
\bottomrule
\end{tabular}}
\label{tab:ttc_ablation}
\end{table}

\paragraph{Impact of temperature parameter $\tau_{\text{temp}}$.}
We further analyze the effect of the temperature parameter $\tau_{\mathrm{temp}}$, which controls the softness of the corruption-aware soft weighting.
As shown in~\cref{fig:tau-temp}, large temperatures ($\tau_{\mathrm{temp}}>0.1$) overly smooth the transition between clean and corrupted regimes, reducing the adaptivity of the counterattack and leading to noticeable drops in both clean and adversarial accuracy.
In contrast, temperatures below $0.1$ produce consistently strong performance, where the soft weighting provides a good balance between sensitivity and smoothness.
Overall, MAC is not highly sensitive to the exact choice of $\tau_{\mathrm{temp}}$ as long as it is not set high, and a moderate value yields stable and near-optimal results across datasets.

\paragraph{Comparisons with state-of-the-art on CLIP ViT-L/14.}
We further evaluate MAC and the compared methods on the strong backbone, CLIP ViT-L/14, across eight fine-grained recognition datasets. 
As shown in \cref{tab:vitl14_finegrained_sota_comparisons}, MAC achieves the highest adversarial robustness on every dataset, reaching an average robust accuracy of \textbf{61.4\%}.
Interestingly, on the Aircraft dataset, MAC even achieves higher adversarial accuracy (31.3\%) than clean accuracy (28.0\%), suggesting that the counterattack can substantially restore corrupted representations and yield exceptional robustness in this domain.
Across all datasets, MAC consistently improves robustness, with gains of up to \textbf{+33.2} points over the best existing tuning-free method.
Furthermore, despite being a \emph{tuning-free} method, MAC surpasses the strongest tuning-based baseline, R-TPT, by a considerable margin (\textbf{+11.8} points on average) while maintaining competitive clean accuracy.
These results demonstrate that MAC delivers state-of-the-art robustness on CLIP ViT-L/14 without requiring any model tuning.

\begin{figure*}[t!]
  \centering
  \includegraphics[width=0.9\linewidth]{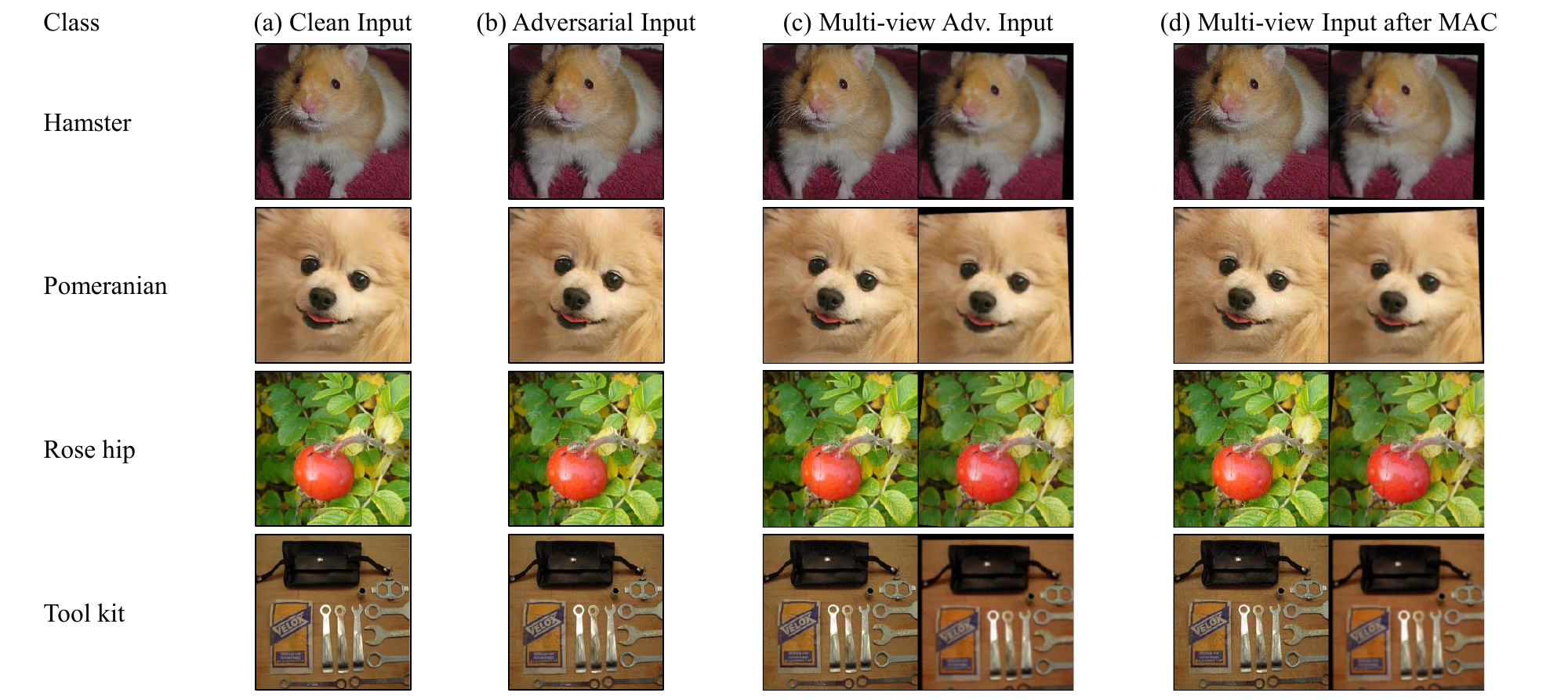}
  \caption{Qualitative visualization of MAC's visual preservation.
  For each example, we show four visual states: the clean image, the adversarial input, the multi-view adversarial inputs produced by our augmentation distribution, and the multi-view adversarial inputs after applying MAC.
  MAC not only preserves the visual quality of the input images but also effectively restores their embeddings toward the clean feature space, as shown in \cref{fig:visual_feats}. This demonstrates that MAC preserves perceptual appearance while mitigating adversarial corruption.}
\label{fig:visual-quality}
\end{figure*}

\paragraph{Ablation study of augmentation components.}
To understand the contribution of each augmentation in MAC, we remove individual components and evaluate the performance, as summarized in \cref{tab:ttc_ablation}.
Removing the affine transform causes a severe robustness collapse from 45.2\% to 6.0\%, showing that geometric variation is essential for generating multi-view inputs and estimating corruption degree.
Other augmentations, such as color jitter, Gaussian noise, and Gaussian blur, provide moderate but meaningful gains, reducing robustness by 0.3 to 2.9 points when removed.
Overall, the full set of augmentations yields the best performance, indicating that diverse geometric and photometric variations jointly strengthen MAC's multi-view guided counterattack and corruption degree estimation.


\paragraph{Visual appearance preservation under MAC}
To qualitatively assess how MAC affects the visual appearance of input images, we examine how the multi-view adversarial inputs are transformed after applying MAC. 
As illustrated in \cref{fig:visual-quality}, we compare four visual states of each sample: the clean image, the adversarial input, the multi-view adversarial inputs generated by our augmentation distribution, and the multi-view adversarial inputs after MAC is applied.
Because PGD perturbations are constrained to be imperceptible, the adversarial input shows no perceptible differences from the clean inputs, even though its embedding is significantly distorted. The multi-view adversarial inputs further include additional augmented variants (\eg, affine transforms, blur, and color jitter) that reflect the diverse views MAC operates on.
After applying our multi-view counterattack with corruption-aware soft weighting, the resulting images remain visually indistinguishable from both the clean and adversarial inputs; no noticeable color inconsistencies or structural distortions are introduced.
Despite this perceptual invariance, the underlying representations are substantially restored toward the clean feature space, as shown in~\cref{fig:visual_feats}.
This demonstrates that MAC effectively mitigates adversarial corruption in the feature space while fully preserving the perceptual quality of the input images.

Since the attacker's configuration is unknown to the defense model, MAC adopts counterattack settings that are independent of the attacker. Although a large $\epsilon^{(\mathrm{ca})}$ does not affect external inputs as MAC's perturbations operate internally within the defense pipeline, we simply choose a conservative $\ell_\infty$ budget of $\epsilon^{(\mathrm{ca})}=8$ with a small number of iterations ($K{=}4$).
This configuration reliably preserves visual quality while still providing strong robustness.

{
    \small
    \bibliographystyle{ieeenat_fullname}
    \bibliography{main}
}


\end{document}